\documentclass[journal]{IEEEtran}
\ifCLASSINFOpdf
   \usepackage[pdftex]{graphicx}
   \graphicspath{{Figure/}{../jpeg/}}
   \DeclareGraphicsExtensions{.pdf,.jpeg,.png}
\else
   \usepackage[dvips]{graphicx}
   \graphicspath{{../eps/}}
   \DeclareGraphicsExtensions{.eps}
\fi
\usepackage{booktabs} 

\usepackage[colorlinks, citecolor=black, urlcolor=blue]{hyperref}
\usepackage{makecell}

\usepackage{amsmath}

\usepackage{amssymb}
 
\usepackage{cite}

\usepackage{algorithm}
\usepackage{algorithmic}

\usepackage{multirow}

\usepackage{diagbox}

\makeatletter
\newcommand{\removelatexerror}{\let\@latex@error\@gobble}
\makeatother

\usepackage{hhline}

\usepackage{bm}

\ifCLASSOPTIONcompsoc
  \usepackage[caption=false,font=normalsize,labelfont=sf,textfont=sf]{subfig}
\else
  \usepackage[caption=false,font=footnotesize]{subfig}
\fi

\usepackage{enumitem}

\usepackage{url}

\begin{document}
%
\title{DiffRoad: Realistic and Diverse Road Scenario Generation for Autonomous Vehicle Testing}

%
%
%

\author{Junjie~Zhou, ~Lin~Wang, ~Qiang~Meng, and~Xiaofan~Wang
\thanks{This work was supported by the National Key R\&D Program of China
with No. 2023YFB4706802, the National Natural Science Foundation of
China with No. 62373245 and 62336005, and in part by the “Dawn” Program
of Shanghai Education Commission, China. The author, Prof. Qiang Meng, would like to appreciate the support of the Ministry of Education of Singapore for this study via the research project MOE-000458-00. \emph{(Corresponding author: Lin Wang.)}}
\thanks{Junjie Zhou and Lin Wang are with the Department of Automation, Shanghai Jiao Tong University, and Key Laboratory of System Control and Information Processing, Ministry of Education of China, Shanghai 200240, China (e-mail: junjie\_zhou@sjtu.edu.cn; wanglin@sjtu.edu.cn).}
\thanks{Xiaofan Wang is with the Department of Automation, Shanghai Jiao Tong University, the School of
Mechatronic Engineering and Automation, Shanghai University, and the School of Electrical and Electronic Engineering, Shanghai Institute of Technology, Shanghai, 201418, China (e-mail: xfwang@sjtu.edu.cn).}
\thanks{Qiang Meng is with the Department of Civil and Environmental Engineering, National University of Singapore, Singapore 117576, Singapore (e-mail: ceemq@nus.edu.sg).}
}

%
%


\markboth{DiffRoad: Realistic and Diverse Road Scenario Library}%
{Shell \MakeLowercase{\textit{et al.}}: Bare Demo of IEEEtran.cls for IEEE Journals}

%




\maketitle


\begin{abstract}
Generating realistic and diverse road scenarios is essential for autonomous vehicle testing and validation. Nevertheless, owing to the complexity and variability of real-world road environments, creating authentic and varied scenarios for intelligent driving testing is challenging. In this paper, we propose DiffRoad, a novel diffusion model designed to produce controllable and high-fidelity 3D road scenarios. DiffRoad leverages the generative capabilities of diffusion models to synthesize road layouts from white noise through an inverse denoising process, preserving real-world spatial features. To enhance the quality of generated scenarios, we design the Road-UNet architecture, optimizing the balance between backbone and skip connections for high-realism scenario generation. Furthermore, we introduce a road scenario evaluation module that screens adequate and reasonable scenarios for intelligent driving testing using two critical metrics: road continuity and road reasonableness. Experimental results on multiple real-world datasets demonstrate DiffRoad's ability to generate realistic and smooth road structures while maintaining the original distribution. Additionally, the generated scenarios can be fully automated into the OpenDRIVE format, facilitating generalized autonomous vehicle simulation testing. DiffRoad provides a rich and diverse scenario library for large-scale autonomous vehicle testing and offers valuable insights for future infrastructure designs that are better suited for autonomous vehicles.
\end{abstract}


%
\IEEEpeerreviewmaketitle

\section{INTRODUCTION}
%
%
%
%

\IEEEPARstart{T}{esting} and validating autonomous vehicles (AVs) necessitate a substantial number of diverse scenarios \cite{feng2023dense}. Road scenarios are the foundation for autonomous driving tests \cite{10056393}. However, there is a significant lack of comprehensive road scenario libraries suitable for digital twin-based simulation testing \cite{feng2021intelligent}. The complexity and variability of real-world road scenarios critically affect the safety and overall performance of AVs. One critical bottleneck in AV simulation testing is the limited availability of diverse road scenarios for simulation purposes, with existing scenarios being predominantly homogeneous. Therefore, there is a pressing need for an effective road scenario generation method that can produce realistic and diverse road scenarios to enhance AV testing.

In addition, the rapid advancement of AV technology is ushering in a transformative era in the transportation sector \cite{9989507}. Unlike human-driven vehicles, AVs operate in fundamentally different ways \cite{shladover2018connected}. However, the existing road infrastructure, primarily developed for human drivers, may not fully capitalize on the benefits provided by this emerging technology \cite{gouda2021automated}. To ensure seamless integration and optimal performance, road infrastructure needs to adapt accordingly.
There is an urgent need for a comprehensive library of road scenarios tailored for digital twin-based AV testing. 
Such a library is essential for guiding the future design of road infrastructure, ensuring it meets the unique requirements of AVs and maximizes their potential benefits.


To achieve high-precision intelligent driving simulation tests, several road scenario formats have been proposed, including OpenDrive \cite{dupuis2010opendrive}, Lanelet2 \cite{poggenhans2018lanelet2}, and CommonRoad \cite{althoff2017commonroad}. Among these, OpenDrive is widely used by most well-known simulators, such as Intel’s CARLA \cite{dosovitskiy2017carla}, Virtual Test Drive (VTD) \cite{von2009virtual}, Apollo \cite{fan2018baidu}, PreScan \cite{Siemens}, and SUMO \cite{krajzewicz2010traffic}. Despite its widespread use, there is little work on generating realistic and diverse road scenarios. Current methods for constructing test road scenarios rely primarily on aerial imagery \cite{azimi2018aerial}, sensor data reconstruction \cite{zheng2019lane}, or manual creation using commercial tools such as MATLAB RoadRunner \cite{MathWorks}. Besides, the CommonRoad Scenario Designer is designed in \cite{maierhofer2021commonroad} to to convert OpenStreetMap (OSM) \cite{bennett2010openstreetmap} maps maps to CommonRoad format. This scenario designer only enables scenario format conversion and cannot generate more scenarios for AV testing.
A model-driven approach is proposed in \cite{zhou2023flyover} to generate interchanges based on the topology graph. However, this method can only generate specific types of interchanges scenarios. In addition, all existing methods are limited to either converting road scene formats or generating a single type of road scene from existing data, and they fall short of generating large-scale, realistic, and diverse road scenes. Therefore, an efficient method for generating a comprehensive library of realistic and diverse road scenarios for AV testing has become a critical challenge.


%

\begin{table*}[!t]
\renewcommand{\arraystretch}{1.2}
\caption{Comparison of existing road scenario generation methods.}
\label{tab1}
\centering
\resizebox{2.06\columnwidth}{!}
{
\begin{tabular}{p{3.2cm}|ccccccccc}
\toprule[1pt]
Method & \makecell[c]{3D\\ Scene} & \makecell[c]{Lane\\ Level} &  OpenDrive & \makecell[c]{Simulation\\ Test}  & Intersection & PUDO & Roundabout & Flyover & Scenes \\ 
\midrule
StreetGAN \cite{hartmann2017streetgan} & - &  -  & -  & - & $\checkmark$ & - & - & - & $\ast$ \\
Conditional GANs \cite{9232519} & - &  -  & -  & $\checkmark$ & $\checkmark$ & - & - & - & $\ast$  \\
CruzWay \cite{9304625} & - & $\checkmark$  & $\checkmark$  & $\checkmark$ & $\checkmark$ & - & - & - & $\ast$  \\
CommonRoad \cite{maierhofer2021commonroad} & - &  $\checkmark$ & $\checkmark$  & $\checkmark$ & $\checkmark$ & - & $\checkmark$ & - & 39,799\\
JunctionArt \cite{muktadir2022procedural} & - &  $\checkmark$  & $\checkmark$  & $\checkmark$ & $\checkmark$ & - & - & - & $\ast$  \\
Predefine \cite{10186533} & - &  $\checkmark$  & $\checkmark$  & $\checkmark$ & -& -& $\checkmark$& - & $\ast$ \\
FLYOVER \cite{zhou2023flyover} & $\checkmark$ &  $\checkmark$  & $\checkmark$  & $\checkmark$ & - & - & - & $\checkmark$ & 1443  \\
GFlowNets \cite{ikram2023probabilistic} & - &  $\checkmark$ & -  & $\checkmark$ & - & - & $\checkmark$ & - & $\ast$  \\
Autoencoder and GAN \cite{yang2023street} & - & -  & -  & - & $\checkmark$ & - & - & - & $\ast$  \\
HDMapGen \cite{mi2021hdmapgen} & - & -  & -  & - & $\checkmark$ & - & - & - & $\ast$ \\
RoBus \cite{li2024robus} & - & -  & $\checkmark$  & $\checkmark$ & $\checkmark$ & - & - & - & 72,400 \\
DriveSceneGen \cite{sun2023drivescenegen} & - & $\checkmark$  & -  & $\checkmark$ & $\checkmark$ & - & - & - & $\ast$ \\
\midrule
DiffRoad & $\checkmark$ &  $\checkmark$  & $\checkmark$  & $\checkmark$ & $\checkmark$ & $\checkmark$ & $\checkmark$ & $\checkmark$ & Infinite \\
\bottomrule[1pt]
\end{tabular}
}
\end{table*}


To address the challenges in road scenario generation for AV testing, we introduce DiffRoad, a diffusion model-based framework that can effectively capture the spatial distribution of road structures and generate high-quality road scenarios. The core idea behind this approach is to perturb the road distribution with noise through a forward diffusion process and then recover the original distribution from Gaussian noise via a learning-based denoising process. This approach results in a highly flexible road scenario generation model.
To achieve controllable road scenario generation, we introduce an innovative road attribute embedding module. This module encodes road attribute information through a wide and deep network, enabling controllable, high-precision road scenario generation. Besides, to further improve the realism and usability of the generated road scenarios, we develop a scene evaluation and screening model. Finally, the evaluated scenarios are automatically converted into a generic OpenDRIVE format, facilitating digital twin-based AV testing.

In summary, our main contributions are as follows:

\begin{itemize}
\item We introduce DiffRoad, the first work to utilize the diffusion model for 3D road scenario generation, pioneering the creation of a diverse lane-level road scenario library. DiffRoad effectively harnesses the diffusion model to capture the distribution of real-world road structures, thereby enabling the generation of realistic and varied road scenarios with high efficiency and transferability.
\item To capture the spatial distribution of road structures and achieve controllable, high-fidelity road scenario generation, we design a novel denoising network architecture called Road-UNet. Road-UNet integrates the FreeU architecture to optimize the balance between the backbone and skip connections, significantly enhancing the quality of generated road scenarios. Furthermore, we introduce a smoothness regularization mechanism to improve the continuity and realism of the generated roads, ensuring smoother transitions and more realistic road layouts.
\item We propose a scene-level scoring function to select the generated road scenarios that are more realistic and sensible. To enhance the generalizability of the generated road scenarios, we automatically convert them into the OpenDRIVE format, making them compatible with mainstream simulation test platforms.
\item The effectiveness of DiffRoad is validated using three road datasets that we collected. The results demonstrate that our method can generate high-fidelity road scenarios while preserving essential statistical properties.
\end{itemize}



\begin{figure*}[t]
\centering
\includegraphics[width=1.0\textwidth]{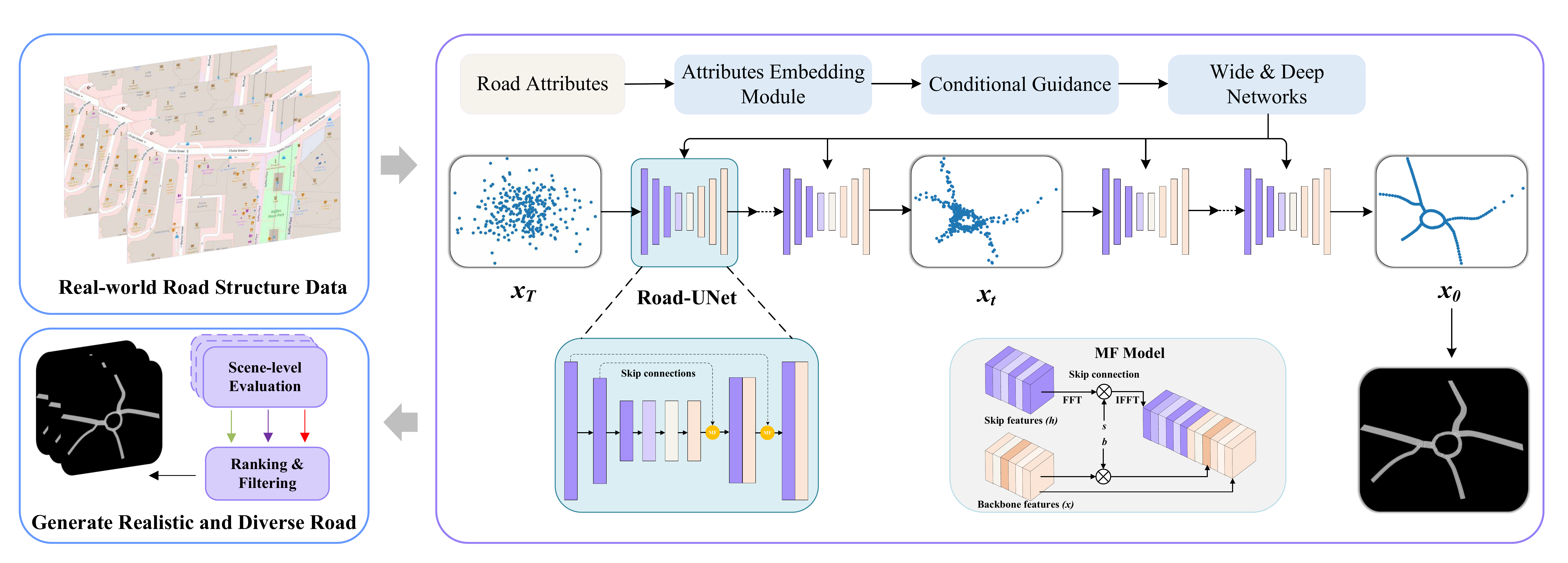}
\caption{The architecture of DiffRoad. The overarching framework of DiffRoad primarily comprises the road structure data generation model based on the enhanced conditional diffusion model,
alongside the scene-level evaluation and filtering module. DiffRoad incorporates the road attention mechanism and Road MultiFreeU-Net (Road-UNet) module, iteratively refining the noise estimation based on the road attribute information to synthesize realistic and diverse road scenarios.}
\label{fig_System structure}
\end{figure*}

\section{Related Work}

\subsection{Road Scenario Generation}

Existing road scenarios for simulation testing are usually created manually using commercial tools or converted from existing data \cite{maierhofer2021commonroad, ZHOU2024123603}. The CommonRoad Scenario Designer \cite{maierhofer2021commonroad} was widely used for converting road scenario formats, including OpenDRIVE, OSM, Lanelet2, and CommonRoad. Additionally, the model-driven method \cite{zhou2023flyover} was proposed for generating a specific type of interchange scenario. Ikram \textit{et al.} \cite{10186533} proposed procedural methods for roundabout generation. However, all these methods are limited to generating specific types of road scenarios, which significantly restricts the scale and diversity of the generated scenario library.

Hartmann \textit{et al.} \cite{hartmann2017streetgan} developed a method that converts road network patches into binary images where pixel intensity indicates street presence or absence. Then, they trained a generative adversarial network (GAN) to synthesize street networks. However, this approach lacks precision and only generates macroscopic road network images. Similarly, Kelvin \textit{et al.} \cite{9232519} used conditional GAN for game map image generation, which proved effective for map image generation. Nevertheless, existing map image generation methods make it difficult to translate into OpenDRIVE road structure data for simulating realistic driving behavior in the virtual simulation software. These methods are limited to macroscopic images and fail to produce accurate micro road structures necessary for AV simulation testing.
Table 1 provides a detailed comparison between DiffRoad and the influential road scenario generation methods.

\subsection{Diffusion Model}

The diffusion model was initially introduced by Sohl-Dickstein \textit{et al.} \cite{sohl2015deep} and later refined by Ho \textit{et al.} \cite{ho2020denoising}.
Over recent years, it has gained recognition as a robust deep probabilistic generative model, achieving state-of-the-art results in numerous applications. These include natural language processing \cite{li2024pre}, image synthesis \cite{dhariwal2021diffusion}, video generation \cite{ho2022imagen}, and molecular design \cite{weiss2023guided}, demonstrating its versatility and effectiveness across a wide range of fields.

In the realm of traffic scenario generation, diffusion models have demonstrated success in trajectory prediction and driving image generation. Chang \textit{et al.} \cite{chang2023controllable} proposed enhanced adversarial optimization objectives into the diffusion model to develop the adversarial agent for AV testing. Additionally, SceneDM \cite{guo2023scenedm} was introduced to jointly generate trajectories of multiple agents using the diffusion model to obtain realistic dynamic driving scenarios. The Versatile Behavior Diffusion model \cite{huang2024versatile} was proposed to achieve trajectory prediction for multiple agents. DriveSceneGen \cite{sun2023drivescenegen} utilized the diffusion model to generate high-fidelity driving scenarios.  While diffusion models have proven effective in generating agent trajectories and driving scenarios, they still call for efforts in spatial road structure generation due to the spatial dependencies and geometric complexities of road structures. To the best of our knowledge, DiffRoad represents a pioneering attempt to employ the diffusion probabilistic model for generating realistic and diverse road scenario libraries.

\section{Methodology}

\subsection{Notions and Preliminaries}

DiffRoad aims to generate realistic and diverse road scenarios for AV testing. In this paper, a road scenario is denoted as a complex set of $n$ interconnected roads, denoted by $\mathbf{x} = \{\mathbf{r}^1, \mathbf{r}^2, \ldots, \mathbf{r}^n\}$, where each road $\mathbf{r}^i = \{m_1^i, m_2^i, \ldots, m_k^i\}$ is a sequence of sampled road shape points. The $j$-th road shape point in the $i$-th road is denoted as a tuple $m_j^i = [lat_j^i, lng_j^i]$, where $lat_j^i$ and $lng_j^i$ indicate latitude and longitude, respectively.


The diffusion probabilistic model utilizes a progressive process wherein noise is incrementally added to the data through a forward diffusion process, subsequently estimating the added noise at each step through a reverse diffusion process, and gradually attenuating the data by removing the noise to recover noise-free data \cite{ho2020denoising}. Essentially, the diffusion probabilistic model embodies a Markovian architecture. 

During the forward diffusion, $T$ time steps of Gaussian noise $N(\cdot)$ are gradually appended to the original data $\mathbf{x}_0 \sim q(\mathbf{x}_0)$ to obtain the latent variables ${\mathbf{x}_1, \ldots, \mathbf{x}_T}$, where $T$ is an adjustable parameter denoting the maximum diffusion step. According to denoising diffusion probabilistic models (DDPMs) \cite{ho2020denoising}, the forward diffusion process is expressed as:
\begin{equation}
\mathbf{x}_t = \sqrt{\overline{\alpha}_t} \mathbf{x}_0 + \sqrt{1-\overline{\alpha}_t} \epsilon, 
\end{equation}
\begin{equation}
\overline{\alpha}_t = \prod_{i=1}^{t} (1- \beta_i), 
\end{equation}
where $\overline{\alpha}_t$ denotes the noise level, $\epsilon$ denotes the noise sampled from the Gaussian distribution $N(0, \mathbf{I})$, and $\beta_i \in (0, 1)$ is the variance schedule.


In the reverse diffusion process, the conditional diffusion model is designed to reconstruct the original road scenario data $\tilde{\mathbf{x}}_0$ from the Gaussian noise data $\tilde{\mathbf{x}}_T \sim N(0, \mathbf{I})$. We design attributes such as road type and road size as conditional information to guide the diffusion model in generating road scenarios that are noise-free and meet specific control requirements. The reverse diffusion process is defined by:
\begin{equation}
p_{\theta} (\tilde{\mathbf{x}}_{0:T} | \mathbf{c}) = p(\tilde{\mathbf{x}}_{T} | \mathbf{c}) \prod_{t=1}^{T} p_{\theta} (\tilde{\mathbf{x}}_{t-1} | \tilde{\mathbf{x}}, t, \mathbf{c}),
\end{equation}
\begin{equation}
p_{\theta} (\tilde{\mathbf{x}}_{t-1} | \tilde{\mathbf{x}}, t, \mathbf{c}) = N(\tilde{\mathbf{x}}_{t-1}; \mu_{\theta} (\tilde{\mathbf{x}}_{t}, t, \mathbf{c}), \Sigma_{\theta} (\tilde{\mathbf{x}}_{t}, t, \mathbf{c})),
\end{equation}
where $\mu_{\theta} (\tilde{\mathbf{x}}_{t}, t, \mathbf{c})$ and $\Sigma_{\theta} (\tilde{\mathbf{x}}_{t}, t, \mathbf{c}))$ are the mean and variance of the reverse process at each step $t$, respectively, $p (\tilde{\mathbf{x}}_{T} | \mathbf{c}) = p (\tilde{\mathbf{x}}_{T}) \sim N(0, \mathbf{I})$, and $\theta$ indicates the parameters of the conditional diffusion model.
Based on previous studies \cite{ho2020denoising}, the mean and variance of the reverse process can be calculated by:
\begin{equation}
\left\{
\begin{aligned}
\mu_{\theta} (\tilde{\mathbf{x}}_{t}, t, \mathbf{c}) & =  \frac{1}{\sqrt{\alpha_t}} (\tilde{\mathbf{x}}_{t} - \frac{\beta_t}{\sqrt{1 - \overline{\alpha}_t}} \epsilon_{\theta}(\tilde{\mathbf{x}}_{t}, t, \mathbf{c})), \\
\Sigma_{\theta} (\tilde{\mathbf{x}}_{t}, t, \mathbf{c}) & =  \sigma_t^2 \mathbf{I}, \text{where } \sigma_t^2 = \left\{
\begin{aligned}
& \frac{1-\overline{\alpha}_{t-1}}{1-\overline{\alpha}_t} \beta_t & t>1, \\
& \beta_1 & t=1,
\end{aligned}
\right.
\end{aligned}
\right.
\end{equation}
where $\epsilon_{\theta}(\tilde{\mathbf{x}}_{t}, t, \mathbf{c})$ indicates the predicted noise under conditions $t$ and road attributes $\mathbf{c}$.

\subsection{DiffRoad Architecture}

The primary goal of DiffRoad is to generate realistic and diverse 3D road scenarios that adhere to specific conditional constraints and preferences. 
The overarching framework of DiffRoad primarily comprises the road structure data generation model based on the enhanced conditional diffusion model, alongside the scene-level evaluation and filtering module, as depicted in Fig. 1. The integration of road structure data encoding and road attribute embedding empowers DiffRoad with comprehensive information, ensuring the controllable generation of road scenarios essential for AV testing. DiffRoad incorporates the road attention mechanism and Road Multi-FreeU-Net (Road-UNet) module, iteratively refining the noise estimation based on real-world road data and corresponding attribute information to synthesize the final road scenarios. Furthermore, we introduce a smoothness regularization term to enhance the overall smoothness of the generated roads.

To address the possibility of generating unrealistic scenarios by the generative model, we introduce a scenario evaluation model aimed at ensuring the usability and realism of the simulation test scenarios. The evaluated road scenarios are subsequently transformed into OpenDRIVE format \cite{dupuis2010opendrive} for AV testing, which has greater compatibility and generalizability for use with a wide range of AV simulation software.

\begin{figure}[tb]
	\label{alg1}
	\renewcommand{\algorithmicrequire}{\textbf{Input:}}
	\renewcommand{\algorithmicensure}{\textbf{Output:}}
	\begin{algorithm}[H]
		\caption{The main processes of DiffRoad}
		\begin{algorithmic}[1]
		\REQUIRE Real-world road structure data $\mathbf{x}$, road attributes $\mathbf{c}$
		\ENSURE Generated road scenes $\tilde{\mathbf{x}}_0$ \\
        \textbf{Training Process:}
        \WHILE{not done}
            \STATE Get conditional guidance $\mathbf{c}$ \\
            \STATE Sample $\mathbf{x}_0 \sim q(\mathbf{x}_0)$ \\
            \STATE Sample $t \sim $ Uniform$(1, \ldots, T)$, $\epsilon \sim N(0, \mathbf{I})$ \\
            \STATE $\mathbf{x}_t = \sqrt{\overline{\alpha}_t} \mathbf{x}_0 + \sqrt{1-\overline{\alpha}_t} \epsilon$ \\
            \STATE $L (\theta) = L_{mse} (\theta) + \omega L_s (\theta)$ \\
            \STATE $\theta = \theta - \eta \nabla_\theta L$
        \ENDWHILE \\
        \textbf{Generating Process:}
        \STATE Get road attributes $\mathbf{c}$ \\
        \STATE Sample $\tilde{\mathbf{x}}_T \sim N(0, \mathbf{I})$ \\
        \FOR{$t \in [T, 1]$}
            \STATE Compute $\mu_{\theta} (\tilde{\mathbf{x}}_{t}, t, \mathbf{c})$ and $\Sigma_{\theta} (\tilde{\mathbf{x}}_{t}, t, \mathbf{c})$ \\
            \STATE Compute the scaling factor $\alpha_z$ and the Fourier mask $\beta_{z, j}$  \\
            \STATE Concatenate augmented skip feature $\mathbf{h}_{z, j}^{'}$ and modified backbone feature $\overline{\mathbf{x}}_{z, j}^{'}$
            \STATE Compute $p_{\theta} (\tilde{\mathbf{x}}_{t-1} | \tilde{\mathbf{x}}, t, \mathbf{c})$
        \ENDFOR 
		\end{algorithmic}
	\end{algorithm}
\end{figure}

\subsection{Road-UNet Architecture}

To ensure accurate prediction of noise at each diffusion time step, the generative model needs to capture the spatial dependencies between road structures. Therefore, we leverage the Road-UNet architecture to construct a network to infer the noise at each diffusion time step in the DiffRoad. 
The Road-UNet comprises two main components: down-sampling and up-sampling, with each layer containing multiple 1D-CNN-based stacked residual network blocks (Resnet blocks). Additionally, a transition module based on the attention mechanism is integrated between these components. 
To enhance noise learning at each diffusion time step, the Road-UNet employs the Sinusoidal embedding method to embed the time step information, which is then fed into each block. Furthermore, Road-UNet incorporates two shared-parameter fully connected layers, which are added to the input of the Resnet block.




To achieve controllable and realistic 3D road scene generation, we meticulously designed the Road-UNet architecture. In contrast to the conventional U-Net model \cite{ronneberger2015u}, we integrate FreeU operations to enhance the quality of road scene generation. Previous research \cite{si2023freeu, zhang2024tc} indicates that backbone features contribute significantly to denoising capabilities, while skip connections enhance high-frequency features in the U-Net model, facilitating noise prediction convergence within the decoder module. However, it will weaken the denoising ability of the backbone. To address this, we introduce Multi-FreeU operations to balance these feature mappings. Specifically, our Road-UNet incorporates two scalar factors, the backbone feature scaling factor $b_z$ for $x_z$ and the skip feature scaling factor $s_z$ for $h_z$, where $x_z$ denotes the backbone feature in the $z$-th block of the Road-UNet decoder, and $h_z$ denotes the skip connection feature in the $z$-th block. For the backbone features $x_z$, we adaptively scale the scaling factor $\alpha_z$ based on the average feature map $\overline{\mathbf{x}}_z$:    
\begin{equation}
\alpha_z = (b_z - 1) \cdot \frac{\overline{\mathbf{x}}_z - Min(\overline{\mathbf{x}}_z)}{Max(\overline{\mathbf{x}}_z) - Min(\overline{\mathbf{x}}_z)} + 1,
\end{equation}
\begin{equation}
\overline{\mathbf{x}}_{z, j}^{'} = \left\{
\begin{aligned}
& \overline{\mathbf{x}}_{z, j} \odot \alpha_z, & \text{if } j < \frac{B}{2}, \\
& \overline{\mathbf{x}}_{z, j}, & \text{otherwise},
\end{aligned}
\right.
\end{equation}
where $b_z$ is the scalar constant, $\overline{\mathbf{x}}_{z, j}$ denotes the $j$-th channel of the feature map $x_z$, $B$ denotes the total number of channels in $x_z$, and $\odot$ denotes element-wise multiplication. To preserve some of the high-frequency details and enhance the realism of the road scene generation, we confine the scaling operation to half of the channels of $x_z$, as described in Equation (7).    

For the skip feature $\mathbf{h}_z$, spectral modulation in the Fourier domain is utilized to selectively attenuate the low-frequency components of the skip features:
\begin{equation}
\mathbf{h}_{z, j}^{'} = IFFT (FFT(\mathbf{h}_{z, j}) \odot \beta_{z, j}),
\end{equation}
\begin{equation}
\beta_{z, j} (r) = \left\{
\begin{aligned}
& s_z, & \text{if } r < r_{thresh}, \\
& 1, & \text{otherwise},
\end{aligned}
\right.
\end{equation}
where FFT$(\cdot)$ and IFFT$(\cdot)$ represent Fourier transform and inverse Fourier transform, respectively. $\beta_{z, j}$ denotes the Fourier mask used to implement the frequency-dependent scaling factor $s_z$. $r$ denotes the radius and $r_{thresh}$ denotes the threshold frequency.

\subsection{Controllable Road Structure Generation}

To generate controllable, realistic, and diverse road scenarios, we propose a novel consistent conditional diffusion method for creating a large number of road scenarios for AV testing. The primary objective of DiffRoad is to approximate the distribution of real-world road scenarios, denoted as \(q(\mathbf{x}_0|c)\), using a parameterized model \(p_\theta(\tilde{\mathbf{x}}_0|c)\). This objective is achieved through a rigorous training and generation process, as detailed in Algorithm 1.


\textbf{Training process:}  
During the training process, the wide and deep networks \cite{cheng2016wide} are initially utilized to encode and embed road attribute information, such as road type and road scale. Subsequently, the DiffRoad model estimates the noise in the reverse process based on the conditional embedding $\mathbf{c}$ and diffusion time step $t$. The objective of training the diffusion model is to minimize the mean square error between the noise $\epsilon_\theta(\mathbf{x}_t, t , \mathbf{c})$ predicted by the model and the Gaussian noise $\epsilon$:
\begin{equation}
L_{mse} (\theta) = \mathbb{E}_{\mathbf{c}, t, \mathbf{x}_0, \epsilon} \left[ \left\| \epsilon - \epsilon_\theta(\mathbf{x}_t, t , \mathbf{c}) \right\|^2 \right].
\end{equation}

To improve the smoothness of the generated road scenarios, we introduce a regularization term into the training objective. This term evaluates smoothness by comparing the distances between adjacent sampling points in the generated scenario with those in real-world road scenarios. Specifically, it ensures that the spacing between points in the generated road scenario mirrors that of actual road data. This smoothness loss term helps align the generated samples with real-world data, minimizing discrepancies and enhancing the fidelity of the generated road scenarios. The smoothness loss term is defined as follows:
\begin{equation}
L_s (\theta) = \mathbb{E}_{\mathbf{c}, t, \mathbf{x}_0, \epsilon} \left[ \left\| (\epsilon^{p+1} - \epsilon^{p}) - (\epsilon_\theta^{p+1}  - \epsilon_\theta^{p} )\right\|^2 \right],
\end{equation}
where $\epsilon^{p+1}$ and $\epsilon^{p}$ represent the $(p+1)$-th point and $p$-th point in the added Gaussian noise $\epsilon$, respectively, $\epsilon_\theta^{p+1}$ and $\epsilon_\theta^{p}$ represent the $(p+1)$-th point and $p$-th point in the predicted noise $\epsilon_\theta$, respectively.

By combining the noise prediction loss and the smoothness loss, we define a novel hybrid objective loss function $L (\theta)$:
\begin{equation}
L (\theta) = L_{mse} (\theta) + \omega L_s (\theta),
\end{equation}
where $\omega$ is the hyperparameter utilized to balance the smoothness loss term ($L_s$) and the mean square error loss ($L_{mse}$) to ensure that $L_s$ does not dominate $L_{mse}$. 

\textbf{Generating process:} 
During the generation process, the DiffRoad model progressively generates realistic and diverse road scenarios from Gaussian noise $\tilde{\mathbf{x}}_T \sim N(0, \mathbf{I})$ guided by the conditional information $\mathbf{c}$. The desired road scenario $\tilde{\mathbf{x}}_0$ is iteratively generated from the noise. By integrating the Road-UNet architecture and adjusting the scaling factors $b_z$ for the backbone features and $h_z$ for the skip features, the details and quality of the generated road scenarios can be further refined.

\begin{figure*}[!t]
\centering
\includegraphics[width=6.6in]{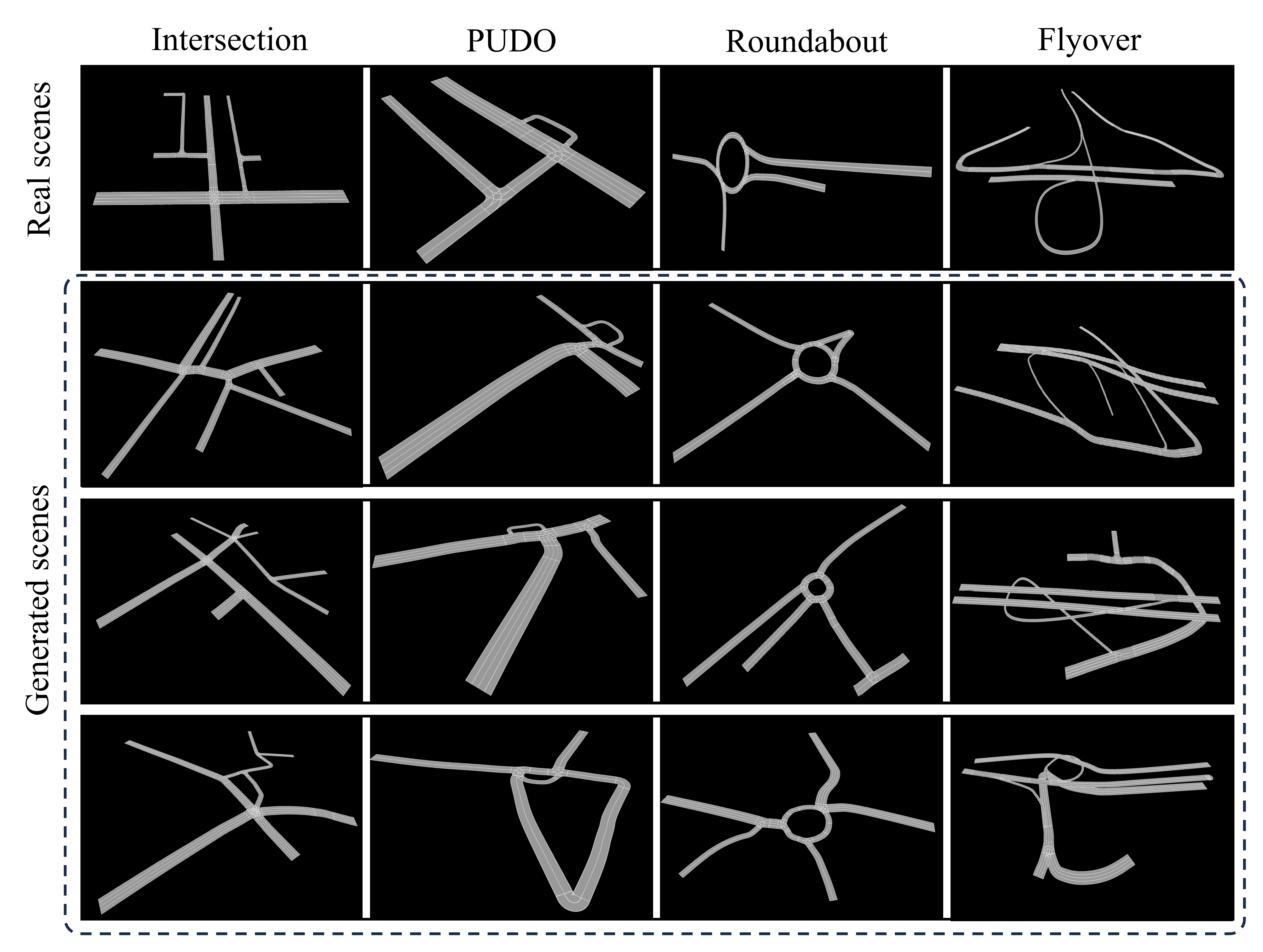}
\caption{Controllable and diverse 3D road scenario generation. DiffRoad enables the generation of specific types of road scenarios based on road attributes and conditional information, thereby meeting the requirements for AV testing.
}
\label{fig_controllable_road_generation}
\end{figure*}

\begin{table*}[t]
\renewcommand{\arraystretch}{1.2}
\caption{Statistics of the collected real-world road datasets.}
\label{STATISTICS}
\centering
\begin{tabular}{ccccc}
\toprule[1pt]
Datasets    \centering & Latitude \centering &  Longitude \centering  & Road Number  &  Average Road Length (m) \\
\midrule
Singapore    \centering & (1.179, 1.478)  &   (103.599, 104.054)  & 26 402 \centering & 1075.227 \\
UK   \centering &  (50.000, 58.000)   &   (-8.000, 2.000)   & 71 484 \centering & 901.561   \\
Japan    \centering &  (34.000, 45.551)  &   (129.000, 153.987)   & 149 561 \centering & 1036.370  \\
\bottomrule[1pt]
\end{tabular}

\end{table*}

\begin{table}[!t]
\renewcommand{\arraystretch}{1.2}
\caption{Hyperparameter setting for DiffRoad.}
\label{tab2}
\centering
\begin{tabular}{lrr}
\toprule[1pt]
Hyperparameter & Setting value &  Refer range \\ \midrule
$T$ (Diffusion Steps) & 500 &  300 $\sim$ 500 \\
Skip Steps & 5 &  1 $\sim$ 10 \\
$\beta$ (linear) & 0.0001 $\sim$ 0.05 &  - \\
$b_z$ (backbone scaling factor) & 1 $\sim$ 1.6 &  1 $\sim$ 1.6 \\
$s_z$ (skip connection scaling factor) & 0.6 $\sim$ 1 &  0.6 $\sim$ 1 \\
Batch size & 1024 &  $\geq$ 256 \\
Sampling blocks & 4 &  $\geq$ 3 \\
Attention blocks & 3 &  $\geq$ 1 \\
Resnet blocks & 2 &  $\geq$ 1 \\
\bottomrule[1pt]
\end{tabular}
\end{table}

\subsection{3D Road Structure Generation}

In this paper, according to the road structure design regulations \cite{hancock2013policy,singapore2019civil,Ministry2017civil},
the maximum longitudinal gradient varies with the design speed. For instance, at 80 km/h, the maximum permissible gradient is 5\%. Vehicles should not be subjected to centrifugal forces $F_c$ exceeding the maximum permissible limits $F_c^{max}$ when navigating up and down ramps. To ensure the safety and comfort of vehicles on ramps, we formulate the following slope calculation formula in combination with the curvature of the road:
\begin{equation}
F_c = \frac{m  v^2 \cos({\rho_c})}{r} \leq F_c^{max},
\end{equation}
\begin{equation}
\rho = \min(\rho_c,\rho_{max}),
\end{equation}
where $\rho$ is the gradient, $m$ is the mass of the vehicle, $v$ is the road design speed limit, and $r$ is the radius of curvature.

\subsection{Scene-level Evaluation Module}

Given that the generative model may produce road scenarios that do not meet specified criteria, we propose a road scene evaluation and screening model to address this issue. This model comprises two key metrics: the road continuity metric and the road reasonableness metric. 
The generated road scenes are screened with reference to the road structure design regulations \cite{hancock2013policy,singapore2019civil,Ministry2017civil}  to obtain compliant road scenarios for AV testing.

Specifically, the road continuity metric is derived by calculating the difference between the mean curvature change rate of the actual road scenario and that of the generated road scenario, denoted as  $w_1(\tilde{\mathbf{x}}_0)$. The road reasonableness metric assesses whether there are overlapping roads in the generated road scenario, quantified by the proportion of overlapping roads to the total number of roads, which is denoted as $w_2(\tilde{\mathbf{x}}_0)$. The final road evaluation function is expressed as follows:
\begin{equation}
S (\tilde{\mathbf{x}}_0) = 100 - (w_1(\tilde{\mathbf{x}}_0) + \lambda w_2(\tilde{\mathbf{x}}_0)),
\end{equation}
where $\lambda$ is the weighting factor used to balance the two terms.

\section{Experiments and Results}

To demonstrate the effectiveness of DiffRoad, extensive experiments were conducted using three real-world road datasets we collected. The experimental results indicate that DiffRoad can generate all types of road scenarios globally for AV testing. 

\subsection{Experimental Setups}

\textbf{Datasets.} We evaluate DiffRoad's generative performance using datasets collected from OpenStreetMap (OSM) \cite{bennett2010openstreetmap}, covering three countries: Singapore, the United Kingdom, and Japan. 
We automatically collected road structure data for major intersections, pick-up and drop-off (PUDO) zones, roundabouts, and flyovers in Singapore, the UK, and Japan from OpenStreetMap \cite{bennett2010openstreetmap} using the OSMnx tool \cite{boeing2017osmnx}. Table 2 provides a summary of the dataset statistics. For each road dataset, we extracted data for twelve roads near the center of each scenario and resampled them to a fixed length using interpolation.

\begin{table*}[!t]

\renewcommand{\arraystretch}{1.2}
\caption{Performance comparison of ablation experiments on Singapore, UK, and Japan datasets.}
\label{table_4}
\centering
\begin{tabular}{c|l|cccc}
\toprule[1pt]
\multirow{2}{*}{Datasets}  & \multirow{2}{*}{Methods}  & \multicolumn{3}{c}{Realism Metric} & \multicolumn{1}{c}{Smoothness} \\  
\cline{3-6}
& & HD ($\downarrow$) & JSD-RL ($\downarrow$)  & JSD-CPD ($\downarrow$) & SISD ($\downarrow$) \\
\midrule
\multirow{7}{*}{Singapore}  & VAE \centering &  138.7514 & 6.4601e-2 & 1.1756e-3 & 47.5050 \\
& GAN \centering &  144.7131 & 5.0690e-2 & 1.6419e-4 & 19.0524 \\
& DiffRoad w/o $\mathbf{c}$ \centering & 145.3339 & 5.8122e-3 & 1.2332e-4 & 36.6263 \\
& DiffRoad w/o Road-UNet  & 141.1844  & 2.5849e-3 & {\bf 2.8907e-5} & 41.5056 \\
& DiffRoad w/o Smooth Loss   & 137.1874 & 9.0295e-3 & 4.9456e-5 & 52.1267 \\
& DiffRoad w/o Scene Evaluation & 136.7471 & 1.7472e-3 & 5.4132e-5 & {\bf 16.1344} \\
& DiffRoad \centering & {\bf 134.2049}  & {\bf 1.6857e-3} & 5.6391e-5 & 18.7556  \\
\midrule
\multirow{7}{*}{UK}  & VAE \centering &  111.5052 & 2.6305e-2 & 2.6226e-4 & 72.1981 \\
& GAN \centering &  114.2630 & 5.5077e-2 & 1.4133e-4 & 48.8026 \\
& DiffRoad w/o $\mathbf{c}$ \centering &  137.6000 & 8.9689e-3 & 8.3407e-5  & 69.9089 \\
& DiffRoad w/o Road-UNet  & 116.7484 & 9.9223e-4 & {\bf 1.5332e-5}  & 46.1778 \\
& DiffRoad w/o Smooth Loss   &  115.2012 & 4.1824e-3 & 1.7703e-5 & 66.0128 \\
& DiffRoad w/o Scene Evaluation &  111.8189 & 7.5581e-4 & 4.2672e-5 & 35.8853 \\
& DiffRoad \centering &  {\bf 110.8399} & {\bf 7.1960e-4} & 4.3191e-5  & {\bf 34.0525 } \\
\midrule
\multirow{7}{*}{Japan}  & VAE \centering &  101.8598 & 2.0482e-2 & 2.5083e-4 & 36.4623 \\
& GAN \centering &  108.0881 & 7.5714e-2 & 1.8674e-4 & 55.3452 \\
& DiffRoad w/o $\mathbf{c}$ \centering &  129.5618 & 2.5187e-2 & 1.6147e-4 & 82.6188 \\
& DiffRoad w/o Road-UNet  &  105.9641 & 1.9669e-3 & 1.5020e-5 & 55.3798 \\
& DiffRoad w/o Smooth Loss   &  103.2421 & 4.5701e-3 & {\bf 9.1420e-6} & 51.9558 \\
& DiffRoad w/o Scene Evaluation &  100.1401 & {\bf  1.4969e-3} & 2.2006e-5 & 34.9363 \\
& DiffRoad \centering &  {\bf 98.8483} & 2.6063e-3 & 2.2471e-5 & {\bf 32.5448} \\
\bottomrule[1pt]
\end{tabular}

\end{table*}

\begin{table*}[t]
\renewcommand{\arraystretch}{1.2}
\caption{Data utility comparison of DiffRoad with existing road scenario libraries.}
\label{table4}
\centering
\begin{tabular}{lccccc}
\toprule[1pt]
Methods  & \makecell[c]{Generation \\ Efficiency \\ (seconds/scenario)}  & \makecell[c]{Intersection\\ (usable/generated)}   & \makecell[c]{PUDO\\ (usable/generated)} & \makecell[c]{Roundabout \\(usable/generated)} & \makecell[c]{Flyover\\ (usable/generated)} \\  
\midrule
CommonRoad & - & 3,901/3,901 & 0  & 1/1 & 0\\
FLYOVER & 300.6 & 0 & 0  & 0 & 1,443/1,443\\
DiffRoad & 0.306 & 420,659/494,894 & 10,980/12,564 & 102,284/113,738 & 65,094/109,082  \\
\bottomrule[1pt]
\end{tabular}

\end{table*}

\begin{figure*}[!t]
\centering
\subfloat[$k$ = 500]{\includegraphics[width=2.2in]{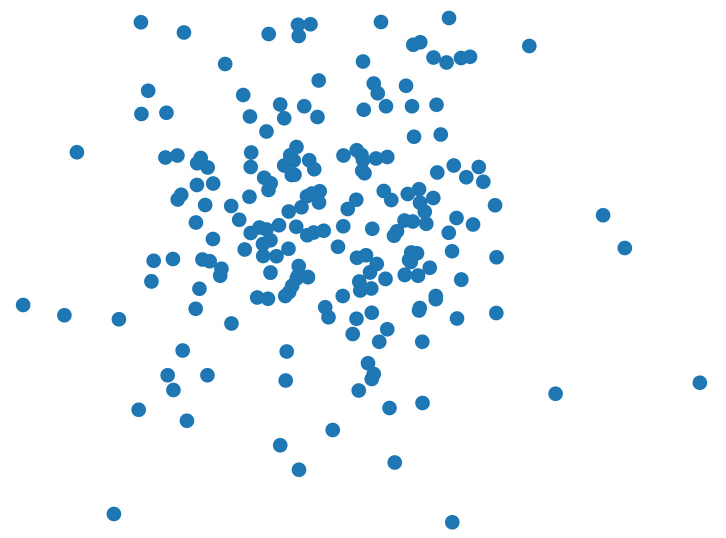}%
\label{fig_pet_NADE_VTD}}
\hfil
\subfloat[$k$ = 100]{\includegraphics[width=2.2in]{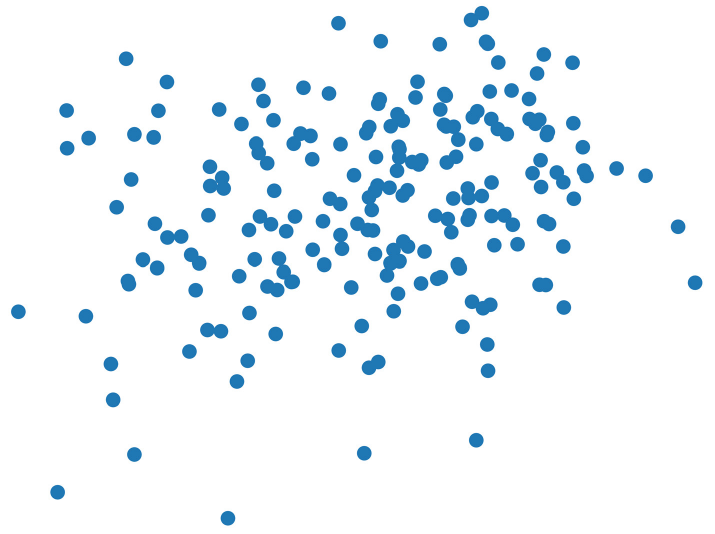}%
\label{fig_Step_100_Sample}}
\hfil
\subfloat[$k$ = 10]{\includegraphics[width=2.2in]{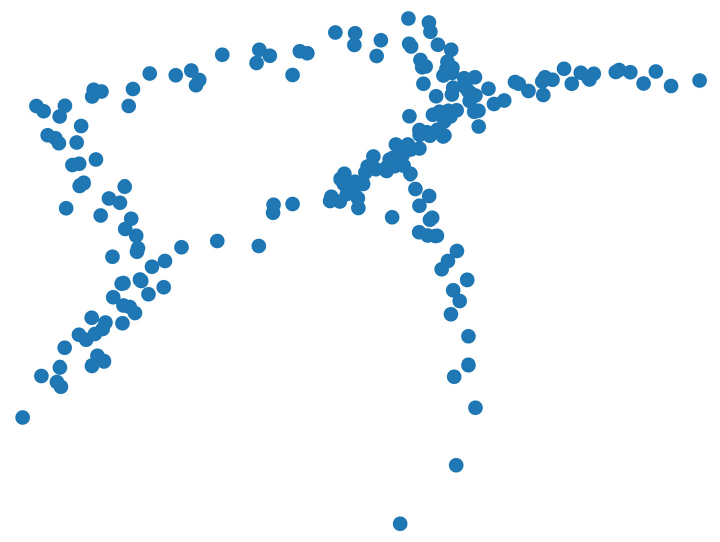}%
\label{fig_Step_10_Sample}}

\subfloat[$k$ = 0]{\includegraphics[width=2.2in]{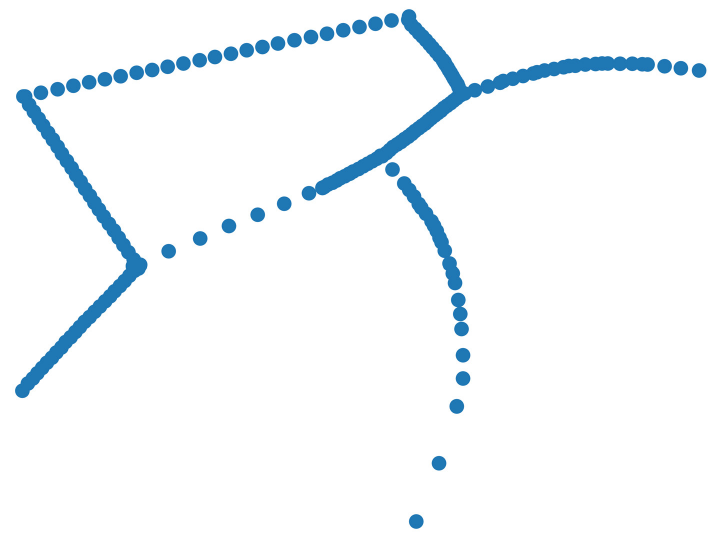}%
\label{fig_Step_0_Sample}}
\hfil
\subfloat[Generated road]{\includegraphics[width=2.2in]{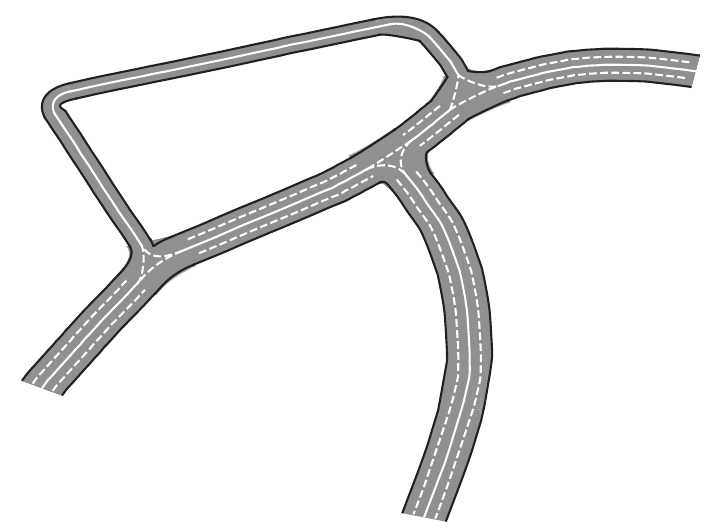}%
\label{fig_generated_road_v0}}
\hfil
\subfloat[Real road]{\includegraphics[width=2.2in]{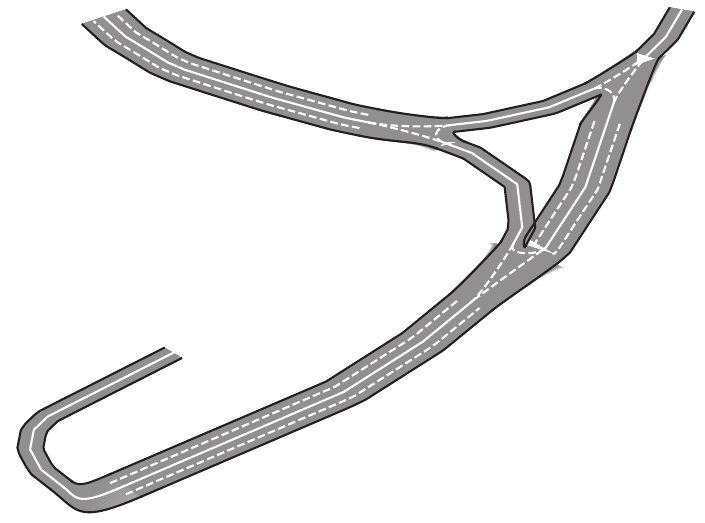}%
\label{fig_real_road}}

\caption{Qualitative results of the DiffRoad model's generation process. The DiffRoad model can generate road scenes based on user input specifying the desired type of scene, such as "three three-way intersections." The model begins by sampling noise from a Gaussian distribution at $k$ = 500. Through a gradual denoising process, DiffRoad refines the sampled noise until convergence at $k$ = 0. The comparative analysis with real-world road scenes reveals that DiffRoad effectively captures the intricate characteristics of real-world road scenes, producing realistic and diverse road configurations.}
\label{fig_Qualitative_result}
\end{figure*}

\begin{figure*}[!t]
\centering
\includegraphics[width=6.6in]{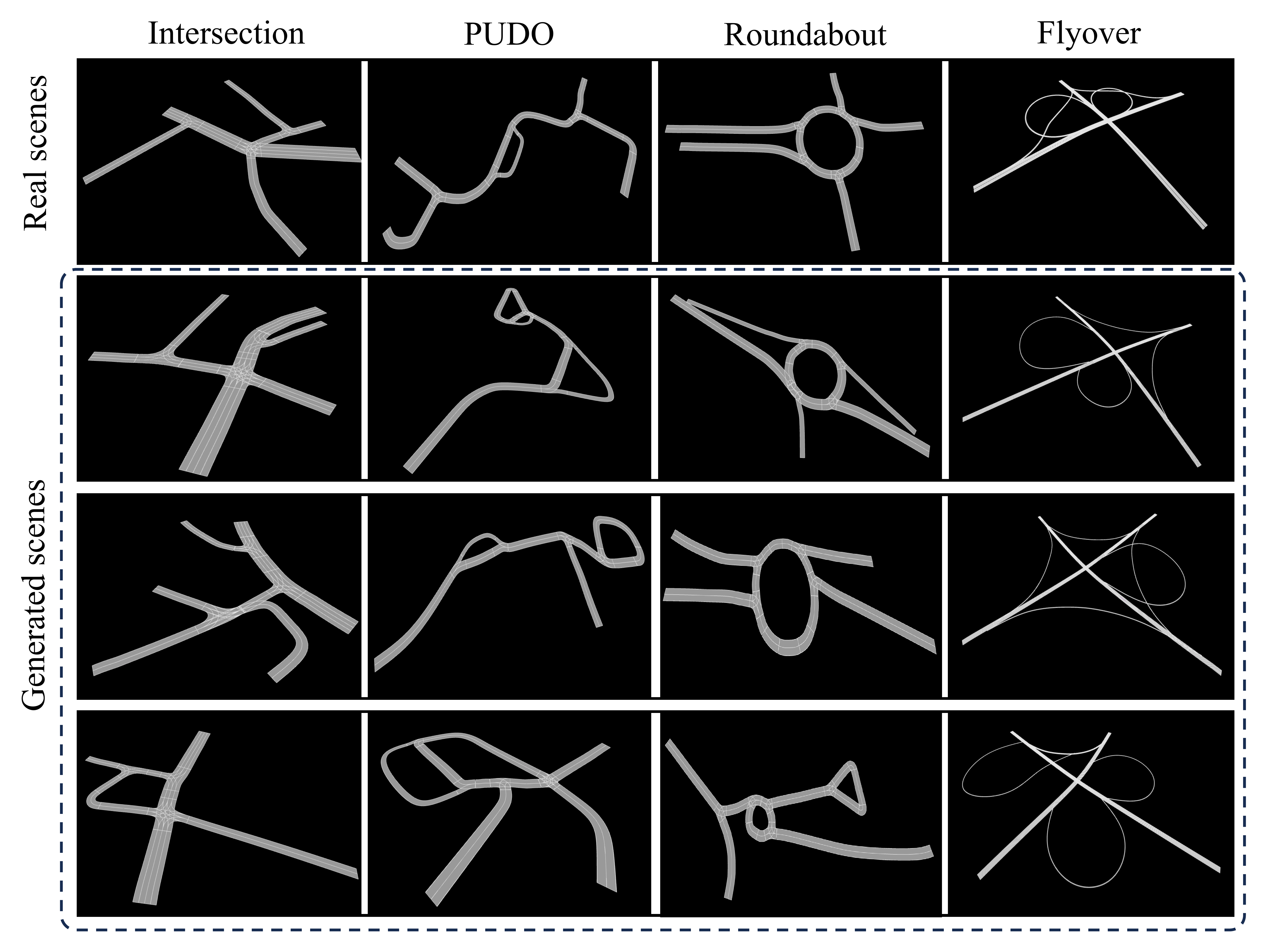}
\caption{Controllable and diverse 3D road scenario generation. DiffRoad enables the generation of specific types of road scenarios based on road attributes and conditional information, thereby meeting the requirements for AV testing.
}
\label{critical_boundary_scenario_fig_more_example1_v1}
\end{figure*}

\begin{figure*}[!t]
\centering
\includegraphics[width=6.6in]{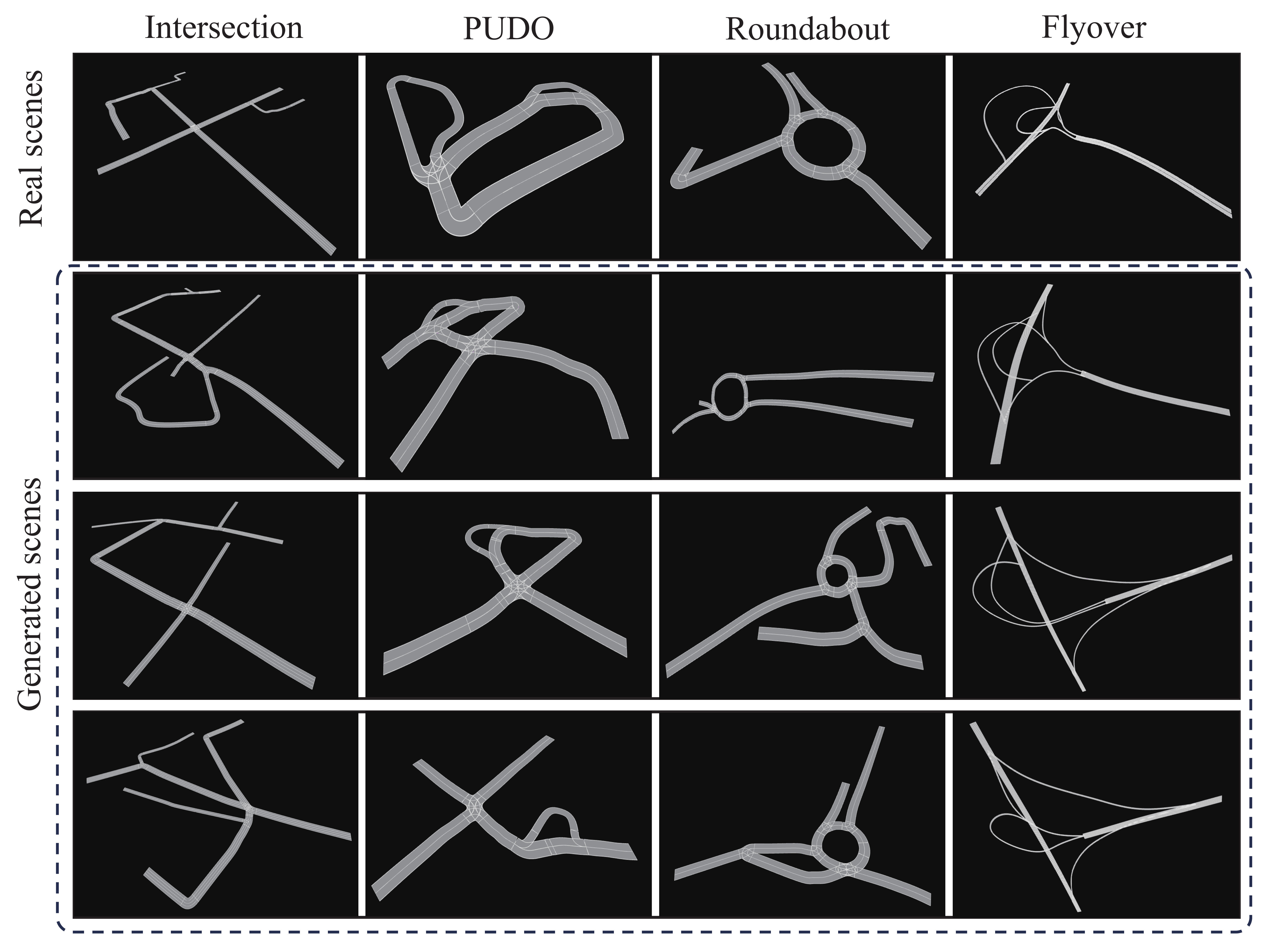}
\caption{Visualization of the generated diverse and realistic 3D road scenarios.
}
\label{fig_generated_road_scenario_fig_v3}
\end{figure*}

\begin{figure}[!t]
\centering
\subfloat[Intersection testing in VTD]{\includegraphics[width=1.6in]{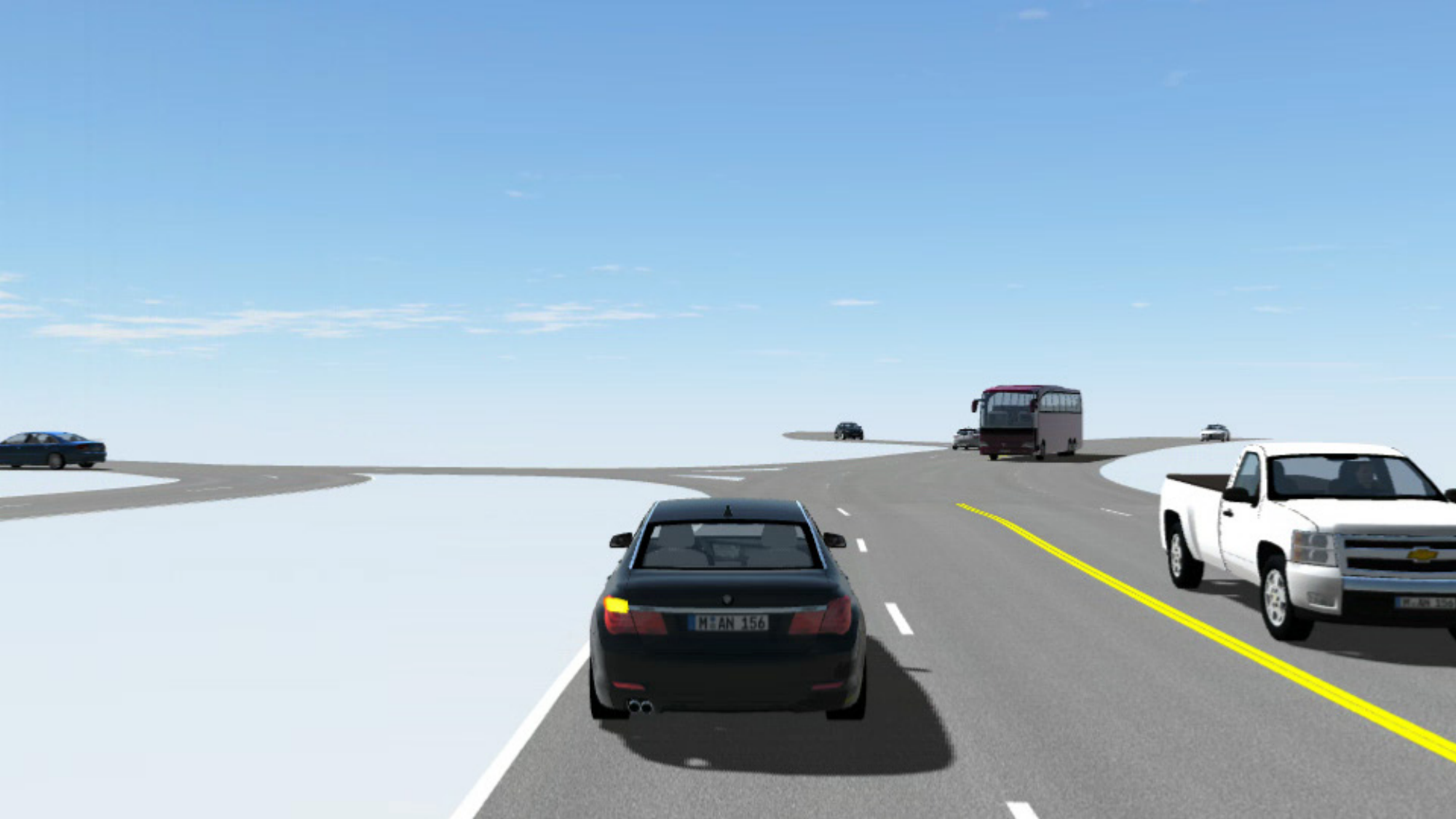}%
\label{fig_Intersection}}
\hfil
\subfloat[Roundabout testing in VTD]{\includegraphics[width=1.6in]{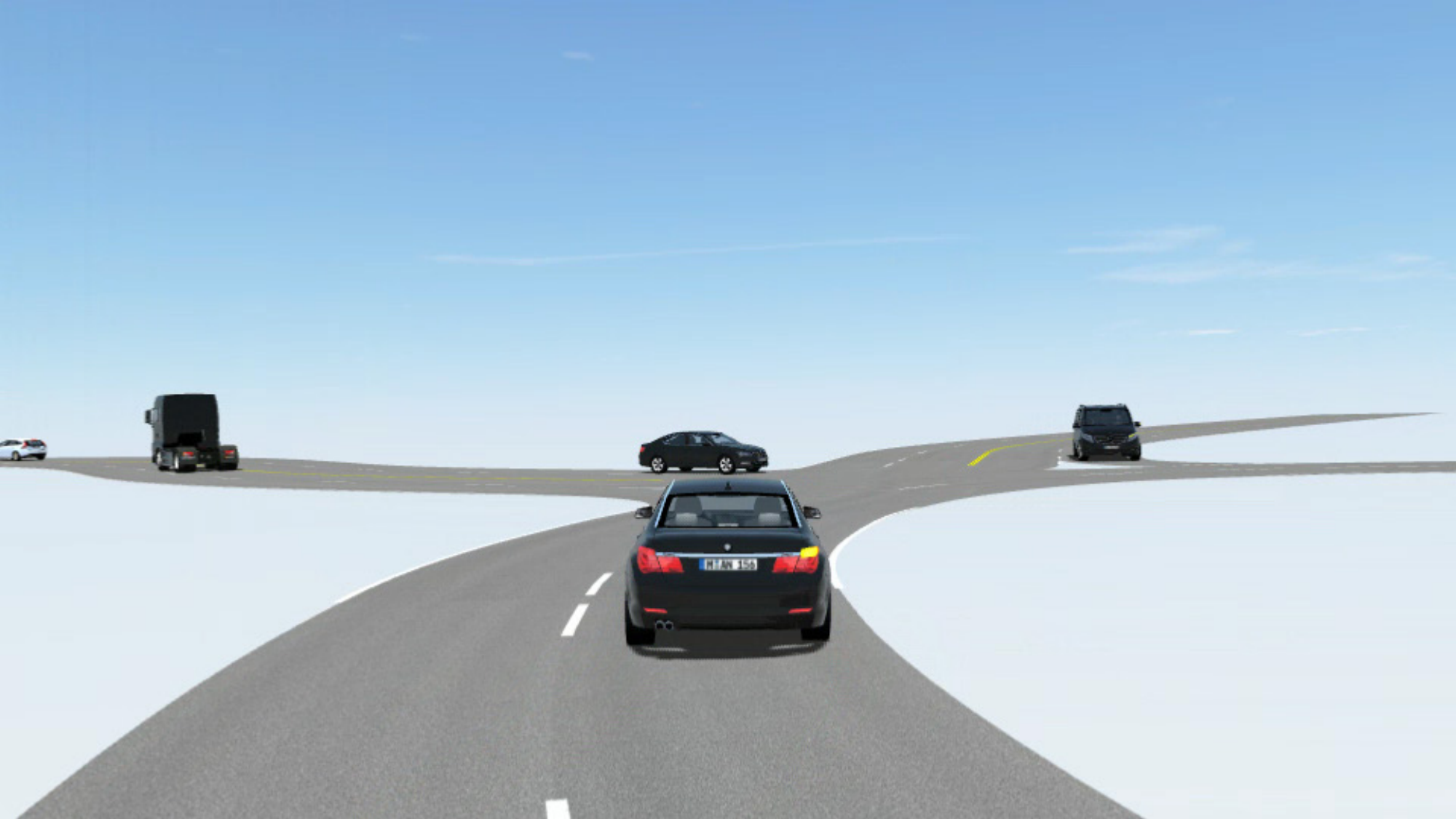}%
\label{fig_PUDO}}

\subfloat[PUDO testing in CARLA]{\includegraphics[width=1.6in]{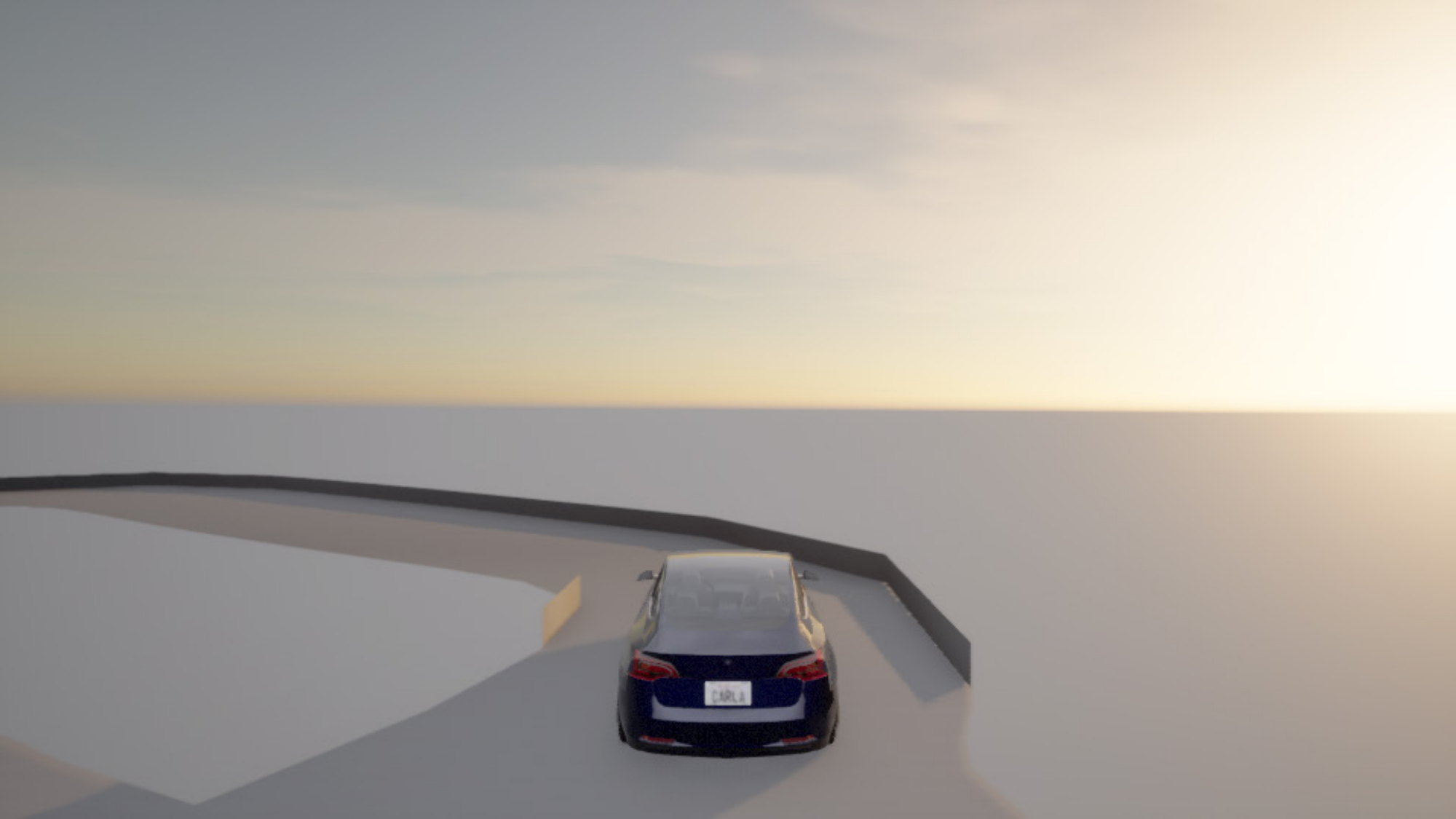}%
\label{fig_Roundabout}}
\hfil
\subfloat[Flyover testing in CARLA]{\includegraphics[width=1.6in]{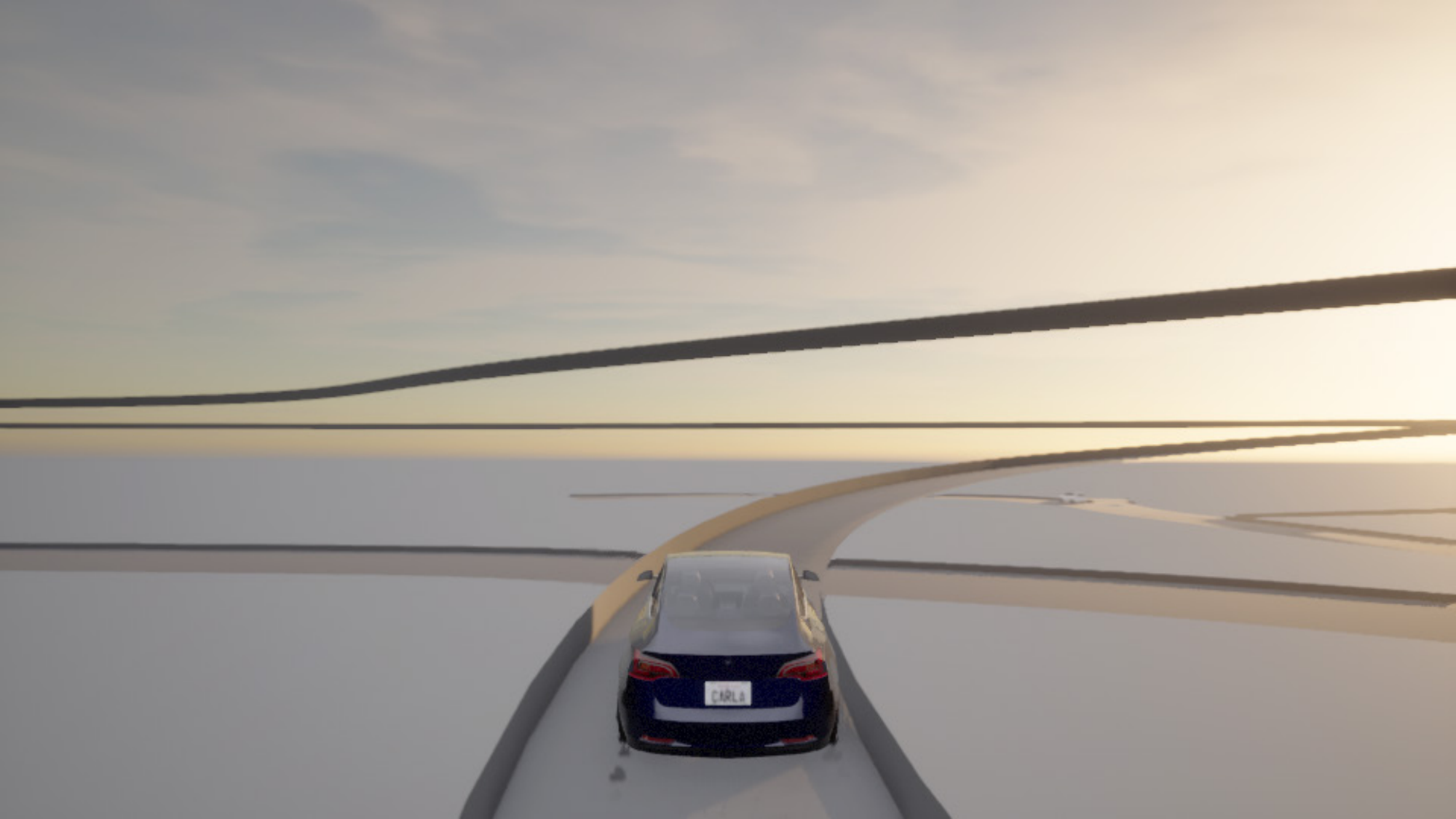}%
\label{fig_Flyover}}

\caption{Visualization results of the realistic and diverse road scenarios generated for intelligent driving simulation tests in different simulation platforms.}
\label{fig_simulation_Utility}
\end{figure}

\textbf{Evaluation Metrics.} The primary goal of road scenario generation is to create realistic and diverse road scenarios for testing AVs. 
To thoroughly evaluate the quality of the generated simulated road scenarios, it is crucial to evaluate their realism and smoothness compared to real-world road scenarios. For realism metrics, we employ the Hausdorff Distance \cite{huttenlocher1993comparing} and Jensen-Shannon Divergence (JSD) \cite{lin1991divergence} to measure the quality of the generated road scenarios. Moreover, for the smoothness metric, we utilize the square integral of the second-order derivative as the measure of smoothness. Specifically, we evaluate the quality of the generated road scenarios with the following metrics:
\begin{itemize}
\item \textbf{Hausdorff Distance (HD):} The Hausdorff Distance is used to compute the similarity between generated road scenarios and real-world road scenarios, providing a measure of the maximum distance from generated road points to the nearest point in real-world road points. Formally, given real-world road data $A$ and generated road data $B$, the Hausdorff distance $d_H(A, B)$ can be calculated by:
\begin{equation}
d_H(A, B) = \max{\{\sup_{a \in A}{\inf_{b \in B}{d(a, b)}}, \sup_{b \in B}{\inf_{a \in A}{d(a, b)}}\}},
\end{equation}
where $d(a, b)$ denotes the Euclidean distance between the road point $a$ and the road point $b$.
\item \textbf{Road Length (JSD-RL):} This metric is designed to evaluate the distribution similarity between the generated road lengths and the real-world road length. We utilize the JSD to quantify the difference in distribution. A smaller JSD value indicates a closer approximation to the real-world statistical characteristics. Let $P$ represent the probability distribution of real-world road lengths, and $Q$ represent the probability distribution of generated road lengths. The JSD can be calculated using the following equation:
\begin{align}
\begin{aligned}
JSD(P, Q) &= \frac{1}{2} D_{KL}(P || \frac{P + Q}{2})  \\
&+ \frac{1}{2} D_{KL}(Q || \frac{P + Q}{2}),
\end{aligned}
\end{align}
where $D_{KL}(\cdot || \cdot)$ is the Kullback-Leibler Divergence \cite{ji2020kullback}.
\item \textbf{Control Point Relative Distance (JSD-CPD):} This metric is designed to evaluate the similarity between the distribution of relative distances among generated road points and the distribution of relative distances among real-world road points. We utilize the JSD to measure the difference between these distributions.
\item \textbf{Road Smoothness (RS):} This metric quantifies the difference in smoothness between real-world roads and generated roads. The smoothness score is derived by calculating the square integral of the second derivative (SISD) of each road.
\end{itemize}

\textbf{Baselines.} As this is the first work to fully automate the generation of large-scale, realistic, and diverse road scenarios for AV testing, we can only further verify the effectiveness and authenticity of the DiffRoad method through comparisons with two typical generative models (VAE \cite{kingma2013auto} and GAN \cite{rao2020lstm}) and a series of ablation studies. Specifically, we compared it against the following variations: 1) DiffRoad w/o $\mathbf{c}$, which examines the contribution of road conditional embedding, 2) DiffRoad w/o Road-UNet, where the conventional U-Net replaces the proposed Road-UNet to evaluate the effectiveness of the Road-UNet structure, 3) DiffRoad w/o Smooth Loss, which examines the contribution of the proposed smoothness loss, and 4) DiffRoad w/o Scene Evaluation, which examines the contribution of the proposed scene evaluation module. 


\textbf{Implementation Details.} For the proposed DiffRoad framework, we summarize the adopted hyperparameters in Table 3. 
All experiments are implemented in PyTorch. All models were trained on 4 RTX4090 using the learning rate of 0.0002. We set the loss weight $\omega$ to 1.0. We set the road evaluation weight $\lambda$ to 1.0.
The training and generating phase of DiffRoad is summarized in Algorithm 1.

\subsection{Overall Generation Performance}

Table 4 presents the performance comparison between DiffRoad and the comparison methods across three real-world datasets. Specifically, we generated 5,000 road scenarios using each generation method and computed all relevant metrics. The results reveal several trends that underscore the advantages of the proposed DiffRoad method.
The conditional version of DiffRoad outperforms the unconditional version (DiffRoad w/o $\mathbf{c}$) because it incorporates road attribute information into the road scenario generation process. This integration enables the model to more effectively consider factors such as road structure type and spatial distribution, which are crucial for accurate road structure generation. This superiority is corroborated by the performance on the Hausdorff Distance metric in Road Realism, where the conditional version significantly outperforms other models.

Furthermore, DiffRoad produces satisfactory results even when the Road-UNet structure, smoothness loss, or scene evaluation module is discarded. Nonetheless, the complete DiffRoad model outperforms its Road-UNet-less variant in terms of both Hausdorff Distance and smoothness metrics. The Road-UNet structure in DiffRoad is specifically designed for spatial structure modeling and multi-scale feature fusion, enabling precise noise level prediction during denoising and thereby generating high-quality road scenarios.

DiffRoad also outperforms the version without smoothness loss on Hausdorff Distance and smoothness metrics. This is because the smoothness loss in DiffRoad is tailored to enhance the smoothness of the road structure, resulting in smoother generated roads. Additionally, the scene evaluation module effectively filters out unreasonable road scenarios, thereby improving the realism and coherence of the final road scenario library.

\subsection{Controllable Road Generation }

Fig. 2 shows the real-world road scenarios and the generated 3D road scenarios. DiffRoad enables the generation of specific types of road scenarios based on road attributes and conditional information, thereby meeting the requirements for AV testing. The first row in the figure shows real-world road scenarios, while the subsequent rows depict various road test scenarios generated by DiffRoad. The first column showcases the intersection scenarios featuring one large intersection and several small three-way intersections generated using condition-specific information. The second column illustrates the pick-up and drop-off (PUDO) scenarios. The third column illustrates the generated roundabout scenarios. Finally, the fourth column displays the flyover scenarios.

As can be seen from the figure, DiffRoad can effectively learn the distribution characteristics of real-world road structures and generate 3D scenarios that closely resemble these structures. Moreover, DiffRoad significantly enhances the diversity of simulation test road scenarios, allowing for a comprehensive evaluation of the performance of AVs across various road types. The integration of controllable condition information enables DiffRoad to accurately generate specified types of 3D road scenarios. This robust capability underscores DiffRoad's versatility in producing meaningful road scenarios tailored to specific conditions.

\subsection{Visualization Results}
We visualize the generated results to better present the performance of DiffRoad. Fig. 3 shows the qualitative results of the DiffRoad model's generation process. The DiffRoad model can generate road scenes based on user input specifying the desired type of scene, such as "three three-way intersections." The model begins by sampling noise from a Gaussian distribution at $k$ = 500. Through a gradual denoising process, DiffRoad refines the sampled noise until convergence at $k$ = 0. The comparative analysis with real-world road scenes reveals that DiffRoad effectively captures the intricate characteristics of real-world road scenes, producing realistic and diverse road configurations.

Fig. 4 and Fig. 5 show the real-world road scenarios and the generated 3D road scenarios. DiffRoad enables the generation of specific types of road scenarios based on road attributes and conditional information, thereby meeting the requirements for AV testing. The first row in the figure shows real-world road scenarios, while the subsequent rows depict various road test scenarios generated by DiffRoad. The first column showcases the intersection scenarios featuring one large intersection and several small three-way intersections generated using condition-specific information. The second column illustrates the pick-up and drop-off (PUDO) scenarios. The third column illustrates the generated roundabout scenarios. Finally, the fourth column displays the flyover scenarios.

As can be seen from the figure, DiffRoad can effectively learn the distribution characteristics of real-world road structures and generate 3D scenarios that closely resemble these structures. Moreover, DiffRoad significantly enhances the diversity of simulation test road scenarios, allowing for a comprehensive evaluation of the performance of AVs across various road types. The integration of controllable condition information enables DiffRoad to accurately generate specified types of 3D road scenarios. This robust capability underscores DiffRoad's versatility in producing meaningful road scenarios tailored to specific conditions.

\begin{figure*}[!t]
\centering
\subfloat[]{\includegraphics[width=1.6in]{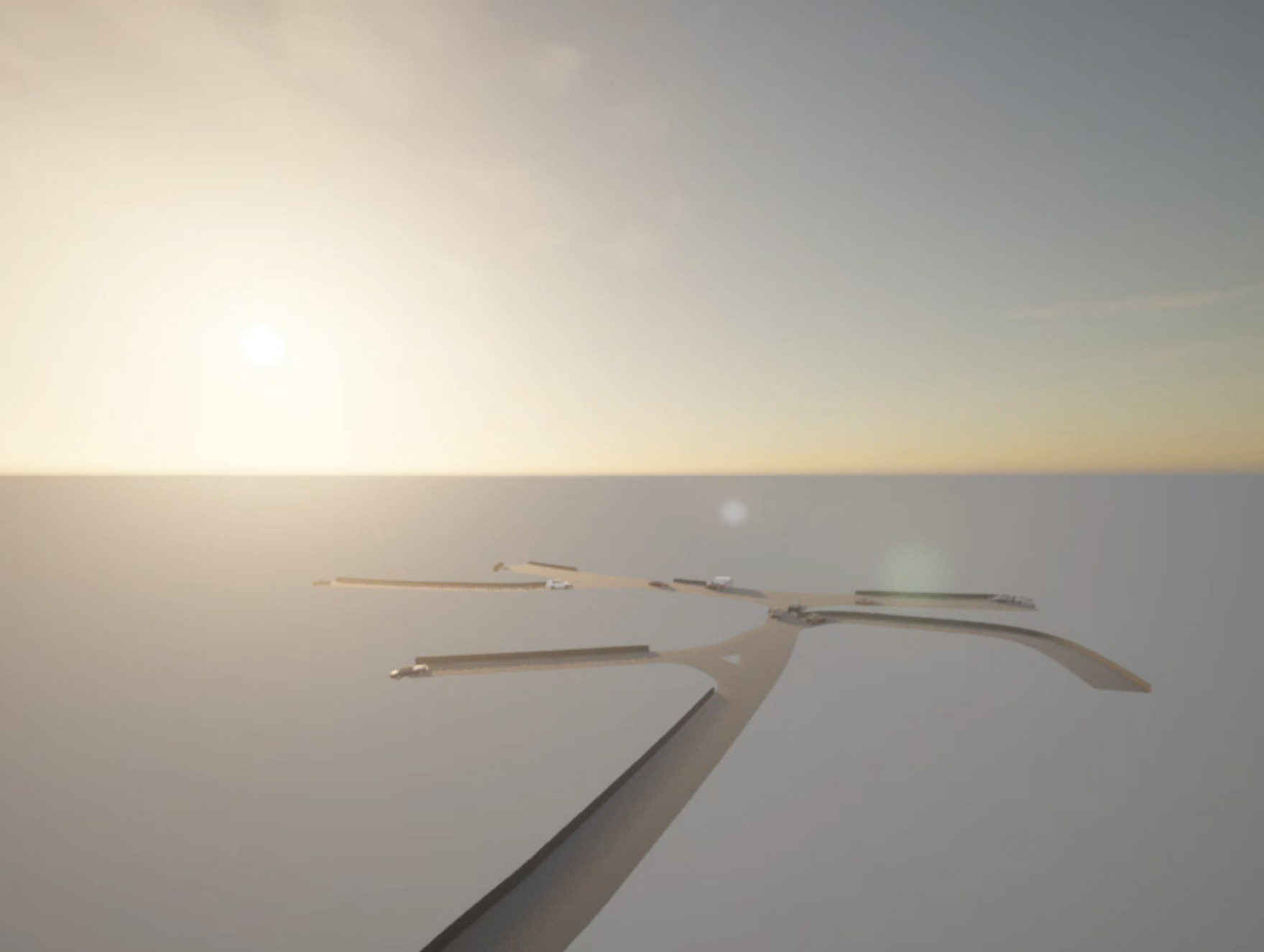}%
\label{fig_Intersection}}
\hfil
\subfloat[]{\includegraphics[width=1.6in]{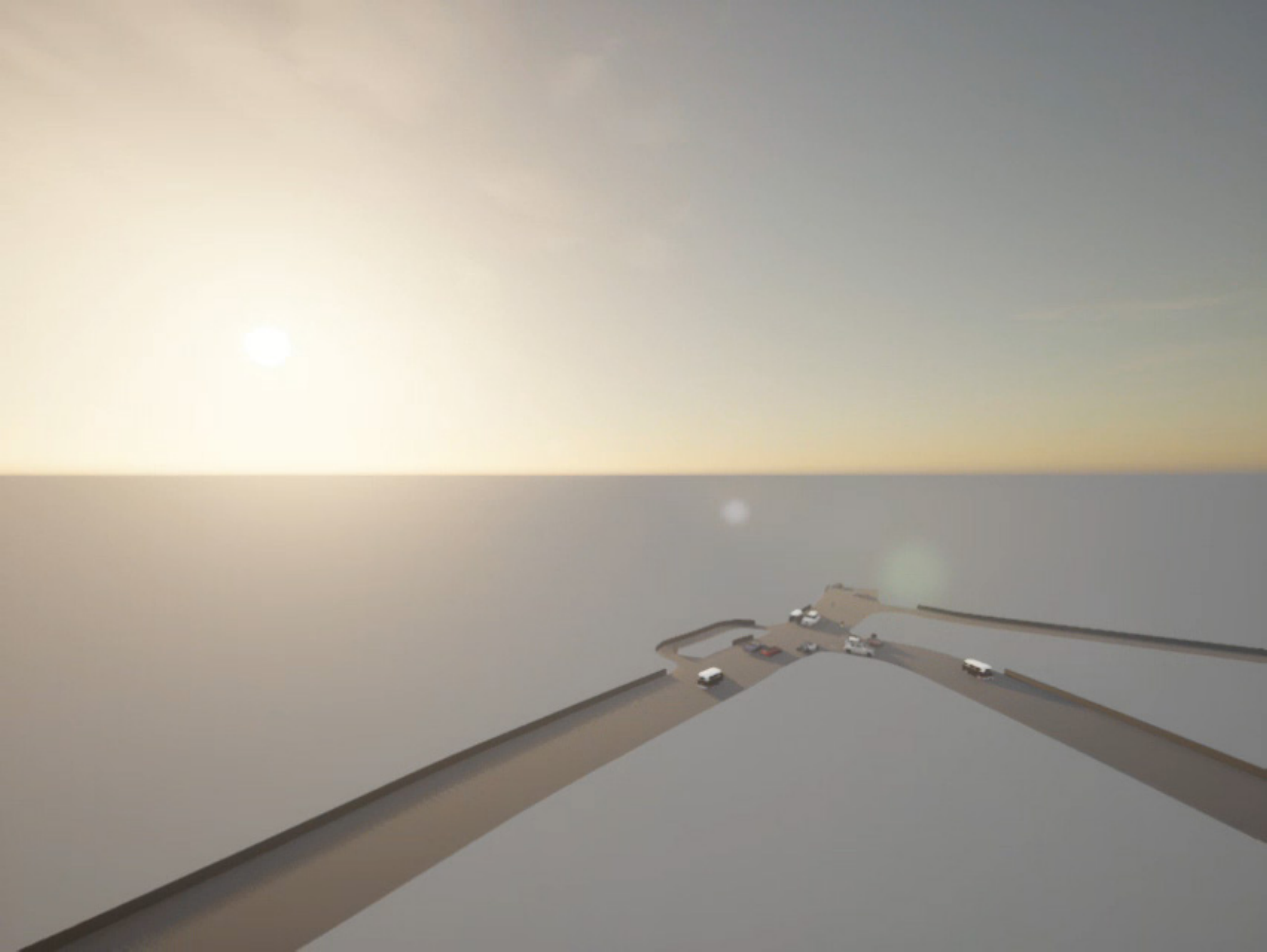}%
\label{fig_PUDO}}
\hfil
\subfloat[]{\includegraphics[width=1.6in]{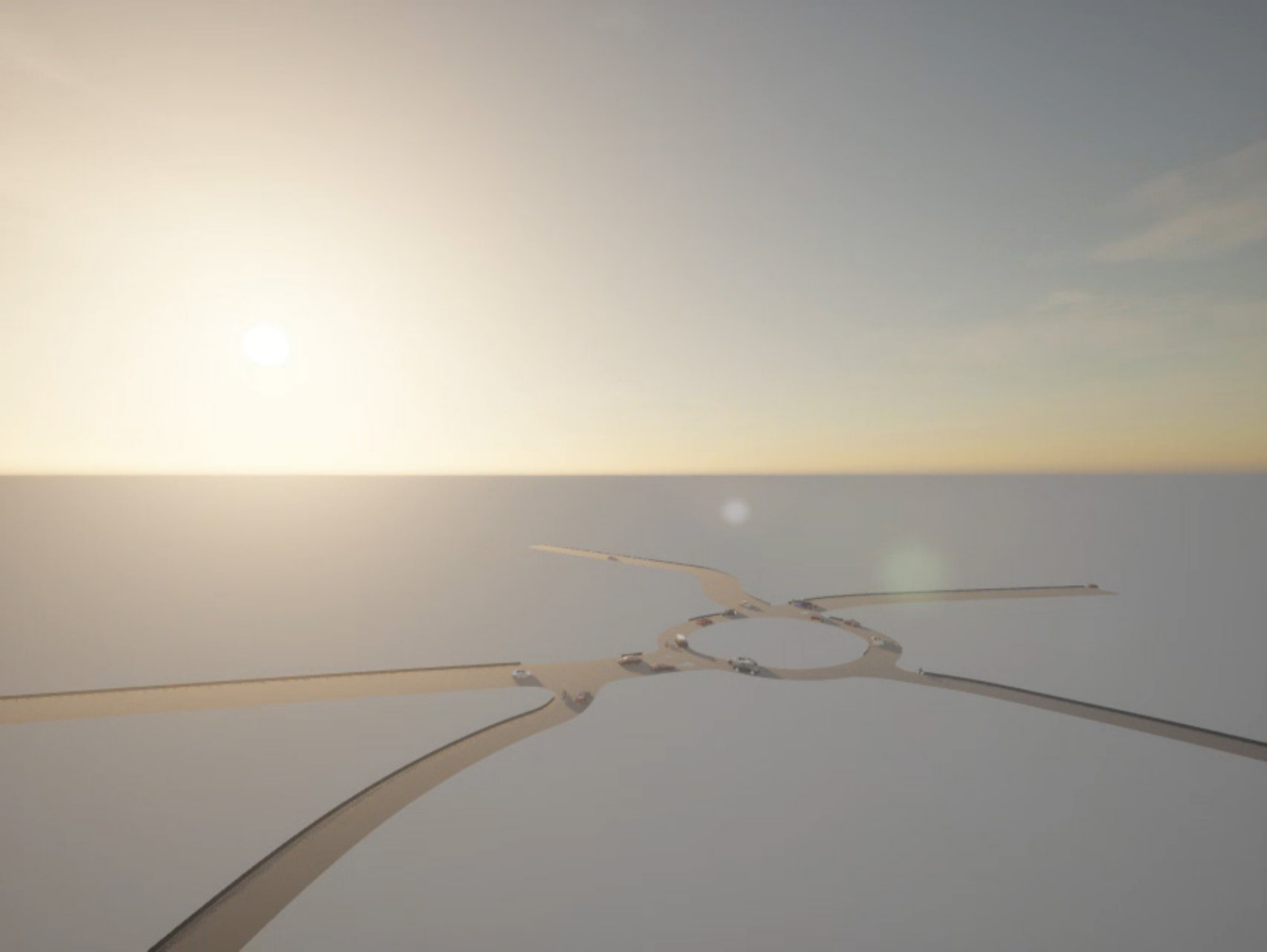}%
\label{fig_Roundabout}}
\hfil
\subfloat[]{\includegraphics[width=1.6in]{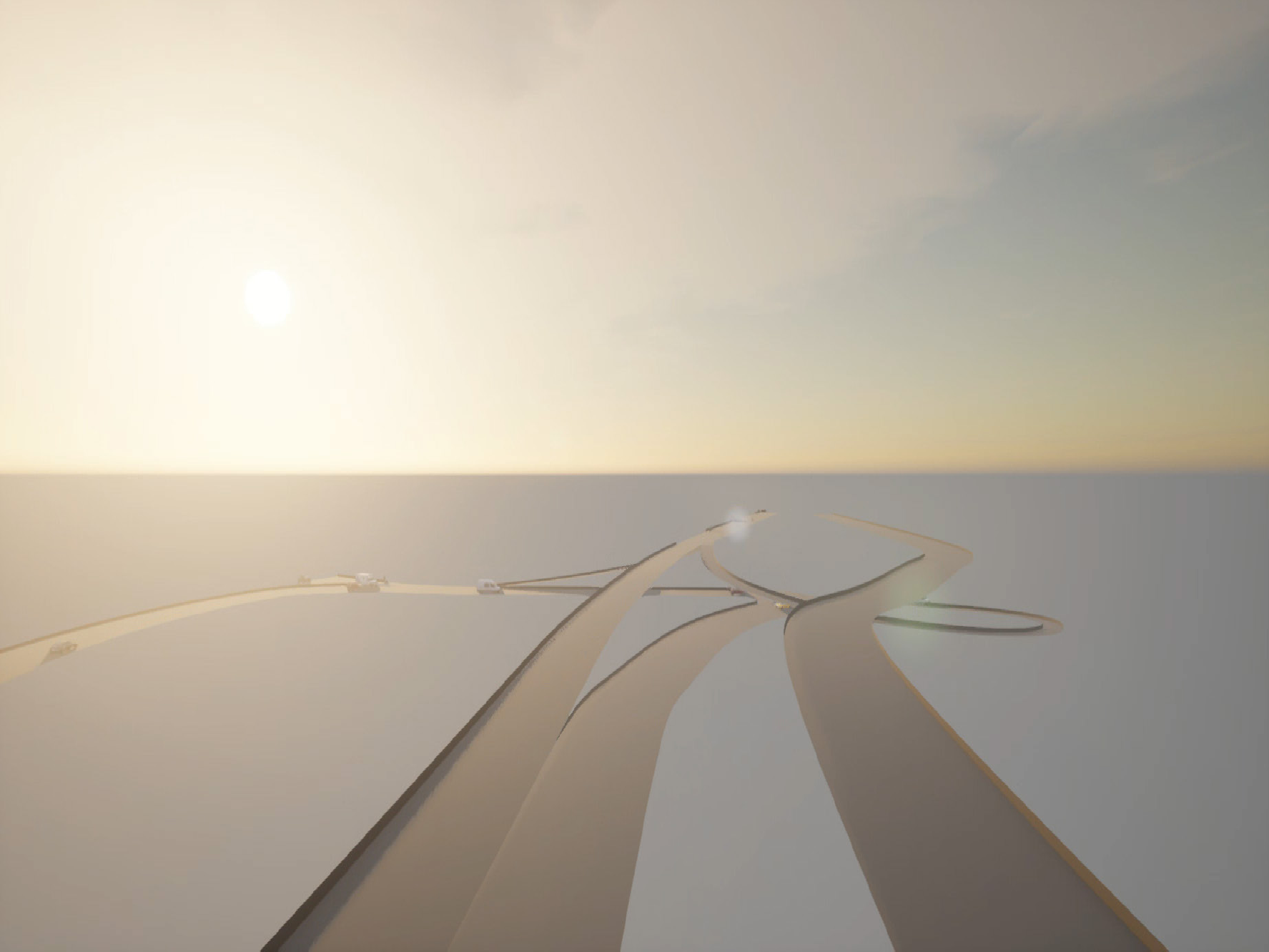}%
\label{fig_Flyover}}

\caption{Global views of the DiffRoad-generated road scenarios imported into the CARLA simulation platform. (a) Simulation test in the intersection scenario. (b) Simulation test in the pick-up and drop-off (PUDO) scenario. (c) Simulation test in the roundabout scenario. (d) Simulation test in the flyover scenario.}
\label{fig_simulation_carla}
\end{figure*}

\begin{figure*}[!t]
\centering
\includegraphics[width=6.6in]{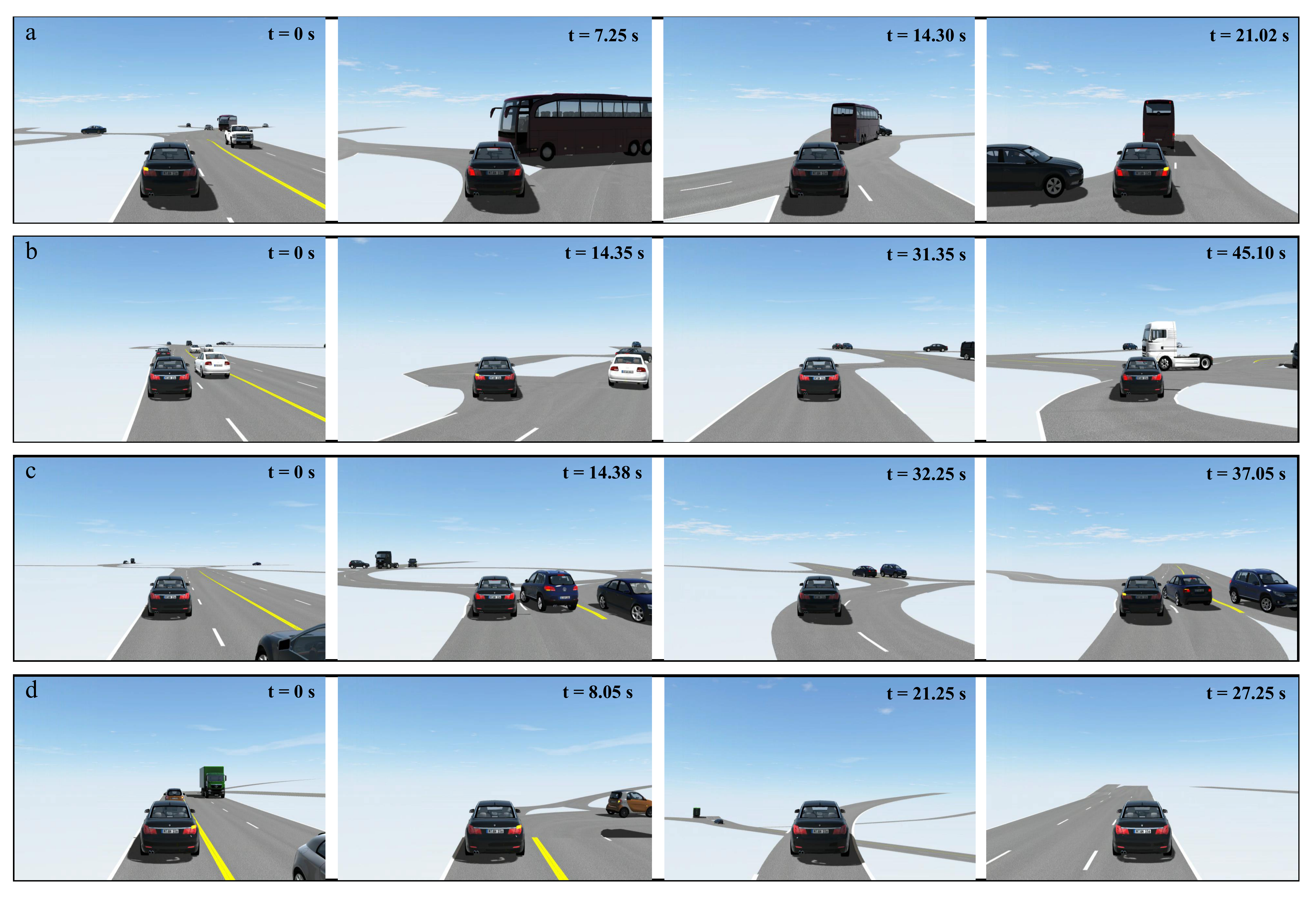}
\caption{Digital twin-based autonomous vehicle
testing experiments using the generated road scenarios on the VTD platform. (a) The VTD AV navigates through various intersections and conflicts with the bus and the car at different intersections. (b) The VTD AV navigates the PUDO scenario, with a trajectory conflict with a truck on the way out of the PUDO. (c) The VTD AV is tested in a roundabout scenario. (d) The VTD AV drives through the flyover scenario.
}
\label{fig_generated_road_scenario_fig_v2}
\end{figure*}

\begin{figure*}[!t]
\centering
\includegraphics[width=6.6in]{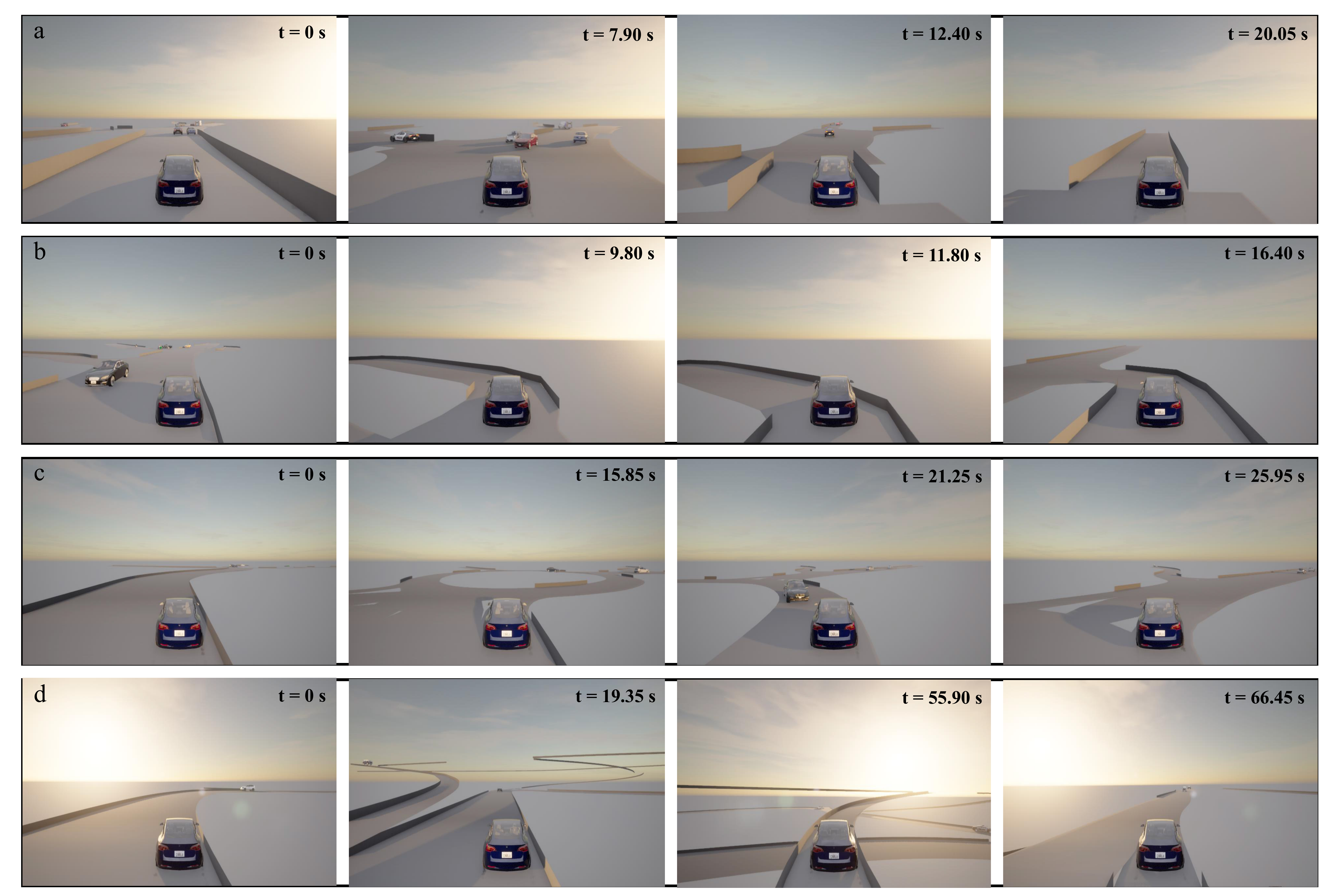}
\caption{Digital twin-based autonomous vehicle
testing experiments using the generated road scenarios on the CARLA platform. (a) The CARLA AV navigates through various intersections. (b) The CARLA AV drives through the PUDO scenario, revealing a collision with the road edge at 11.8 seconds. (c) The CARLA AV is tested in a roundabout scenario. (d) The CARLA AV drives through the flyover scenario.
}
\label{generated_road_scenario_fig_v2}
\end{figure*}

\subsection{Utility of Generated Road}

The generated road scenario library effectively overcomes the limitations of existing simulation road scenarios, which are often monotonous and manually constructed, leading to inefficiencies. As concluded in Table 5, DiffRoad significantly boosts efficiency, achieving a 982 times improvement over the optimized FLYOVER method. Fig. 6 illustrates the generalizability and effectiveness of scenarios generated for intelligent driving testing, including intersections, roundabouts, PUDO zones, and flyovers. The results demonstrate that DiffRoad's outputs are both versatile and applicable across various simulation platforms, showcasing strong generalization and scalability.

To further validate the effectiveness of the DiffRoad-generated 3D road scenarios for autonomous vehicle testing, we conducted additional digital twin-based autonomous vehicle testing experiments using the generated road scenarios. Intelligent driving test experiments were performed using the CARLA simulation platform \cite{dosovitskiy2017carla} and VTD simulation platform \cite{von2009virtual} to demonstrate the generalizability of the scenarios generated by DiffRoad. The OpenDRIVE \cite{dupuis2010opendrive} scenarios generated by DiffRoad were imported into both CARLA and VTD, serving as the test environments for autonomous vehicles. To demonstrate the generalizability and effectiveness of the DiffRoad-generated road scenarios, we employed generic autonomous driving models on both CARLA and VTD platforms for evaluation.

Fig. 7 provides an overview of the road scenarios generated by DiffRoad within the CARLA simulation platform. Specifically, Fig. 7(a) depicts the intersection scenario, Fig. 7(b) displays the pick-up and drop-off (PUDO) scenario, Fig. 7(c) illustrates the roundabout scenario, and Fig. 7(d) depicts the flyover scenario. These snapshots highlight DiffRoad’s capability to create diverse 3D road structures suitable for AV testing.

Fig. 8 illustrates the results of the intelligent driving simulations on the VTD platform for left-hand traffic. In Fig. 8(a), the VTD AV navigates through various intersections and conflicts with the bus and the car at different intersections. Fig. 8(b) illustrates the AV navigating the PUDO scenario, with a trajectory conflict with a truck on the way out of the PUDO. Fig. 8(c) shows the AV is tested in a roundabout scenario, conflicting with surrounding vehicles as it enters and exits the roundabout. Fig. 8(d) illustrates the AV passing through the flyover scenario.

Fig. 9 presents the visualization results of intelligent driving simulations on the CARLA platform for right-hand traffic. In Fig. 9(a), the CARLA self-driving vehicle navigates through multiple intersections. Fig. 9(b) illustrates the AV navigating the PUDO scenario, revealing a collision with the road edge at 11.8 seconds. This also proves that the DiffRoad-generated scenarios can effectively test the safety of AVs across various road shapes. Fig. 9(c) shows the AV is tested in a roundabout scenario, while Fig. 9(d) illustrates the AV passing through the flyover scenario.

These experimental results demonstrate that DiffRoad-generated scenarios effectively assess AV safety under diverse road conditions. Besides, these results indicate that the road scenarios generated by DiffRoad can be effectively utilized for AV testing, overcoming the limitations of current methods in quantity and variety. Unlike existing studies that can only generate road images, DiffRoad produces road structure data that is automatically converted into 3D road scenarios in OpenDRIVE format for intelligent driving tests. Additionally, the proposed method exhibits strong generalization and scalability, enabling the generation of any type of road scenario and compatibility with various simulation test platforms. The generated comprehensive scenario library significantly accelerates AV testing and lays the groundwork for future performance improvements in AV technology.

\section{Conclusion}

To achieve realistic and diverse 3D road scene generation for AV testing, we propose DiffRoad, a novel architecture based on the improved diffusion model that automates the entire process by learning the data distribution of real-world road structures. To ensure controllability, we design a road attribute information embedding module that guides the generation of specific types of road scenes. To capture real-world road structural features and enhance generation quality, we develop the Road-UNet architecture, incorporating the FreeU network for more accurate noise level prediction. Furthermore, we introduce a scene-level evaluation module to filter and select the most realistic road scenes for AV testing. The generated road scenes are automatically converted into the generic OpenDRIVE format, ensuring compatibility with existing mainstream autonomous driving simulation software. Extensive experiments demonstrate that DiffRoad can generate realistic, diverse, and controllable road scenes that align with the statistical characteristics of real-world road scenes. Further experiments validate the effective application of DiffRoad-generated road scenes in digital twin-based intelligent driving simulation tests. In the future, it is worth exploring the use of the large-scale road scenario library generated by
DiffRoad to further accelerate intelligent driving testing and investigate road infrastructure that is more suitable for AVs.

\ifCLASSOPTIONcaptionsoff
  \newpage
\fi




\bibliographystyle{IEEEtran}
\bibliography{main}

\end{document}